\PassOptionsToPackage{prologue,dvipsnames}{xcolor}
\documentclass[10pt,twocolumn,letterpaper]{article}

\usepackage{multirow}
\usepackage{tikz}
\usepackage{pgfplots}
\pgfplotsset{compat=newest}
\usepgfplotslibrary{units}
\usepackage{graphicx}
\usepackage{subcaption}

\usepackage{amsmath}  % 如果还没有添加的话
\usepackage{colortbl}  % 在导言区添加

\usepackage{iccv}              % To produce the CAMERA-READY version
\definecolor{iccvblue}{rgb}{0.21,0.49,0.74}
\usepackage[pagebackref,breaklinks,colorlinks,allcolors=iccvblue]{hyperref}

\def\paperID{9203} % *** Enter the Paper ID here
\def\confName{ICCV}
\def\confYear{2025}

\title{Revisiting Pool-based Prompt Learning for Few-shot Class-incremental Learning}

\author{
  Yongwei Jiang \quad
  Yixiong Zou\thanks{Corresponding author.} \quad
  Yuhua Li \quad 
  Ruixuan Li\\
  \textnormal{School of Computer Science and Technology, Huazhong University of Science and Technology}
  \\
  \texttt{\small \{jiangyongwei, yixiongz, idcliyuhua, rxli\}@hust.edu.cn}
  \\
}

\begin{document}
\maketitle
\begin{abstract}

Few-Shot Class-Incremental Learning (FSCIL) faces dual challenges of data scarcity and incremental learning in real-world scenarios. While pool-based prompting methods have demonstrated success in traditional incremental learning, their effectiveness in FSCIL settings remains unexplored. This paper presents the first study of current prompt pool methods in FSCIL tasks, revealing an unanticipated performance degradation in incremental sessions. Through comprehensive analysis, we identify that this phenomenon stems from token-dimension saturation: with limited data, excessive prompts compete for task-relevant information, leading to model overfitting. Based on this finding, we propose LGSP-Prompt (Local-Global Spatial Prompting), which innovatively shifts pool-based prompt learning from the token dimension to the spatial dimension. LGSP-Prompt generates spatial prompts by synergistically combining local spatial features and global frequency-domain representations to highlight key patterns in input images. We construct two spatial prompt pools enabling dynamic prompt selection to maintain acquired knowledge while effectively learning novel sessions. Extensive experiments demonstrate that our approach achieves state-of-the-art performance across multiple FSCIL benchmarks, showing significant advantages in both base knowledge preservation and incremental learning. Our implementation is available at \href{https://github.com/Jywsuperman/LGSP}{https://github.com/Jywsuperman/LGSP}.

\end{abstract}    
\section{Introduction}
\label{sec:intro}

\begin{figure}[t]
    \centering
    \includegraphics[width=\columnwidth]{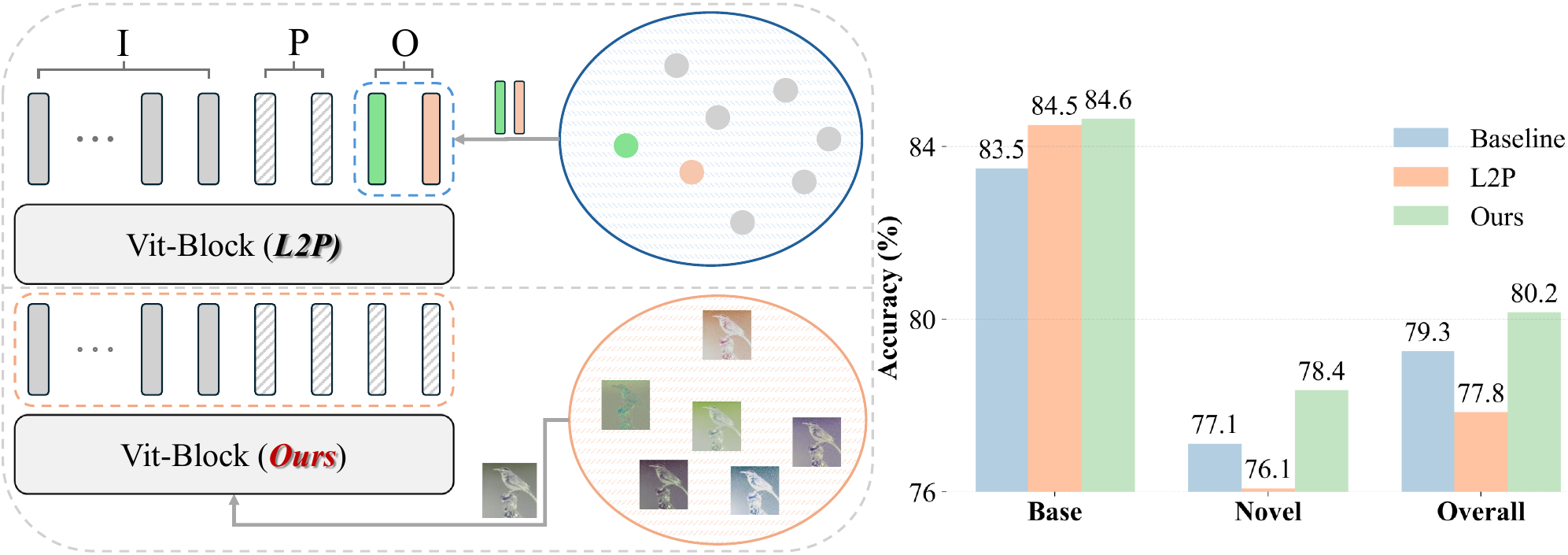}\vspace{0.3cm}
    \caption{
    %Visualization of our key innovation in addressing token-dimension saturation for FSCIL. 
    Traditional pool-based methods (L2P) stack multiple prompts in the token dimension, which shows promising results in balancing the incremental-session learning and the catastrophic forgetting.    
    However, when applying it to the FSCIL task, we find an unanticipated performance degradation on incremental sessions of the FSCIL task.
    In this paper, we delve into this phenomenon and find the degradation stems from the token-dimension saturation problem, which inspires us to shift the pool-based prompt learning from the token dimension to the spatial dimension and effectively address this problem.
    %leading to performance degradation in novel sessions due to prompts competing for limited task-relevant features. Our LGSP-Prompt (bottom) innovatively explores the spatial dimension for prompt learning, where input tokens ($I$) and VPT prompts ($P$) are enhanced by spatially-adaptive prompts ($O$) that synergistically combine local spatial features and global frequency-domain representations. Right: 
    %Experimental validation shows that our spatial prompting approach consistently outperforms both the baseline and L2P methods across all metrics, demonstrating superior performance in base knowledge preservation, novel class adaptation, and overall incremental learning.
    }
    \vspace{0.4cm}
    \label{fig:method_comparison}
\end{figure}

In real-world scenarios, data typically arrives incrementally with limited samples, presenting significant challenges for model adaptation. Such practical constraints have given rise to Few-Shot Class-Incremental Learning (FSCIL)~\cite{tao2020fscil}, which requires models to learn from scarce samples in novel sessions while maintaining previously acquired knowledge in the base session.

To address this challenge, recent works have been conducted through the adaptation of pre-trained models. Visual Prompt Tuning (VPT)~\cite{jia2022visualprompttuning} has emerged as a promising approach to enable efficient model adaptation while preserving rich pre-trained knowledge, which makes it a strong baseline for ViT-based FSCIL~\cite{park2024pretrainedvisionlanguagetransformers}.
Based on it, the pool-based prompt learning mechanism, pioneered by Learning to Prompt (L2P)~\cite{wang2022learning} and further enhanced by DualPrompt~\cite{wang2022dualprompt}, presents an elegant solution through dynamic prompt maintenance and selection for different incremental tasks. However, although these approaches have demonstrated success in traditional incremental learning scenarios, their effectiveness in FSCIL scenarios remains unexplored. 

In this paper, we revisit such a pool-based prompt learning mechanism for the FSCIL task.
Surprisingly, by integrating the prompt pool with current VPT-based FSCIL methods, we observe a notable degradation of novel-session performance (Figure~\ref{fig:method_comparison}), albeit a slight performance improvement on the base session, in contrast to the traditional incremental learning scenarios.
This phenomenon consistently manifests across different experimental settings, indicating a fundamental challenge in the direct application of prompt pool methods to FSCIL scenarios.

To understand this phenomenon, we conduct comprehensive investigations. 
Our analysis reveals that the performance degradation stems from limited training data in novel sessions, 
where scarce training data cannot allow the model to take too many prompts to improve the performance. 
To verify this finding, we then visualize the attention map and identify two critical effects as the number of selected prompts increases: (1) a global attention shift, where the model's focus gradually deviates from discriminative foreground features to task-irrelevant background elements, and 
(2) local interference between different prompts' attention patterns, where prompts compete for task-relevant patterns and losers majorly focus on task-irrelevant ones. Through this systematic analysis, we identify that the fundamental challenge lies in the prompts' token-dimension saturation: Given limited training samples, the model does not need so many prompts as input tokens to contain the information from training samples, i.e., saturation. Therefore, prompts compete with each other to encode useful information, while the loser can only encode the useless ones, leading to overfitting and degrading the CLS token's representation.

This finding inspires us to rethink an alternative perspective: Given that VPT is already inserted in the ViT as the strong baseline method, rather than adding the prompt pool in the saturated token dimension, we explore the spatial dimension~\cite{liu2024insvp, kirillov2023segment} as a new aspect for pool-based prompting of the model. 
Building on this intuition, we introduce LGSP-Prompt (Local-Global Spatial Prompting), a novel framework that circumvents token-dimension saturation through dual-perspective spatial adaptation. 
Our approach synergistically combines local spatial features for fine-grained pattern recognition and global frequency-domain representations for holistic understanding, which outputs a spatial prompt to highlight certain spatial patterns in the input image (Fig.~\ref{fig:method_comparison}). 
Based on it, we build two pools of spatial prompts, enabling dynamic prompt selection to maintain acquired knowledge and learn the novel sessions.

Extensive experiments demonstrate that our approach significantly advances the state-of-the-art in FSCIL across multiple benchmarks, showing consistent improvements in both base knowledge preservation and incremental learning. 

In summary, our key contributions include:

\begin{itemize}
    \item To the best of our knowledge, we are the first to study the pool-based prompt learning methods for the FSCIL task.
    
    \item Our analysis is the first to reveal the token-dimension saturation in pool-based learning, which is the root cause of performance degradation in the few-shot scenarios.
    
    \item Based on the analysis, we propose a novel pool-based spatial prompt learning framework that leverages both local and global spatial information while avoiding token-dimension saturation.
    
    \item Extensive experiments validate our state-of-the-art performance across multiple FSCIL benchmarks.
\end{itemize}
\section{Related Work}
\label{sec:related}

\subsection{Visual Prompting and Pool-based Extensions}
Inspired by prompt learning in NLP~\cite{brown2020language,liu2021pre,lester2021power}, Visual Prompt Tuning (VPT)~\cite{jia2022visualprompttuning} introduces learnable prompt tokens into vision transformers while keeping the backbone frozen. This approach has shown performance comparable to fine-tuning while modifying minimal parameters. The field has evolved in multiple directions: efficient prompt design~\cite{cheng2023e2vpt,gao2022visualprompttuningtesttime,zhang2025pvitprioraugmentedvisiontransformer} reduces computational cost through token pruning, instance-level adaptation~\cite{liu2024insvp,huang2023diversityawaremetavisualprompting,dong2023lptlongtailedprompttuning} explores dual-level prompting for better features, and prompt pool mechanisms~\cite{wang2022learning,wang2022dualprompt,wang2023codaprompt,wang2023spromptslearningpretrainedtransformers,wang2023hierarchicaldecompositionpromptbasedcontinual} maintain dynamic prompts for different tasks. While these pool-based methods succeed in traditional continual learning, their application in FSCIL faces unique prompt-information competition challenges.

% 审稿人的意见：
% While these pool-based methods succeed in traditional continual learning, their application in few-shot class-incremental learning faces unique challenges related to token-dimension saturation and prompt competition under data scarcity.

\subsection{Few-Shot Class-Incremental Learning}
Few-Shot Class-Incremental Learning (FSCIL) addresses learning from limited samples while maintaining model performance on previous tasks~\cite{akyürek2022subspaceregularizersfewshotclass,tao2020fscil,zou2024flattenlongrangelosslandscapes,zou2024a,zou2024attention}. Recent works explore various directions: structure-based approaches~\cite{tao2020fscil,Zhang_2021_CVPR,dynamicBackbone,Ahmad_2022_CVPR,kang2023softsubnetworkfewshotclassincremental} focus on dynamic network architectures for adaptive feature extraction, knowledge preservation strategies~\cite{kukleva2021lcwof,rajasegaran2020self,10203568CABD,cui2023uncertainty,Shankarampeta2021FewShotCI} employ distillation and memory replay for continual learning, while feature optimization~\cite{peng2022fewshotclassincrementallearningopenset,zhou2022forward,zhou2023few,kim2023warping,wang2022fosterfeatureboostingcompression,PAN2024110686} and meta-learning approaches~\cite{zhu2021selfpromotedprototyperefinementfewshot,zou2022marginbasedfewshotclassincrementallearning,zou2024compositionalfewshotclassincrementallearning} explore innovative strategies for few-shot scenarios. Recent pre-trained model-based methods~\cite{radford2021learningtransferablevisualmodels,dosovitskiy2021imageworth16x16words,liu2024fewshotclassincrementallearning,tang2024rethinking,Wang2024OnTA,park2024pretrainedvisionlanguagetransformers,huang2024learningpromptdistributionbasedfeature,ma2024reconstructiontargetmattersmasked,han2023decouplingrepresentationknowledgefewshot} establish new paradigms for effective feature alignment and transfer. However, existing methods often require complex architectural modifications or sophisticated optimization strategies, while the potential of prompt-based learning remains largely unexplored in FSCIL.

%-------------------------------------------------------------------------
%-------------------------------------------------------------------------
\section{Understanding Token-Dimension Saturation in Pool-based Prompt Learning}
\label{sec:analysis}

\subsection{Preliminaries}
\label{sec:preli}

Few-Shot Class-Incremental Learning (FSCIL) spans $T$+1 sessions, beginning with a base session followed by $T$ novel sessions. In the base session, the model learns from a large-scale dataset $\mathcal{D}^0$ containing abundant samples across $|\mathcal{C}^0|$ classes. Subsequently, in each novel session $t$ $(1 \leq t \leq T)$, the model is trained on a small set of new classes $\mathcal{D}^t$ organized in an $N$-way $K$-shot format, where only $K$ training samples are available for each of the $N$ novel classes. Importantly, the class sets across different sessions are mutually exclusive ($\mathcal{C}^i \cap \mathcal{C}^j = \emptyset, i \neq j$), and at each session $t$, only the current session's data $\mathcal{D}^t$ is accessible for training. The model's performance is evaluated on the expanding test set $\mathcal{D}_{test}^{t} = \cup_{i=0}^t \mathcal{D}_{test}^{(i)}$, which encompasses all previously encountered classes, requiring the model to maintain knowledge of base classes while adapting to novel ones with minimal samples.
%-------------------------------------------------------------------------
%-------------------------------------------------------------------------
\subsection{Novel-Session Performance Degradation due to Limited Training Samples}
%-------------------------------------------------------------------------
Recent advances in prompt learning have shown promising results in preserving model knowledge while enabling efficient adaptation in FSCIL. Building upon it, prompt pool methods have further demonstrated strong capabilities in preventing catastrophic forgetting through dynamic prompt selection and maintenance. Motivated by these successes, a natural direction is to integrate prompt pool mechanisms with Visual Prompt Tuning (VPT) for FSCIL. 

%-------------------------------------------------------------------------
% 从pool中选出prompts length增长时，base+novel性能变化
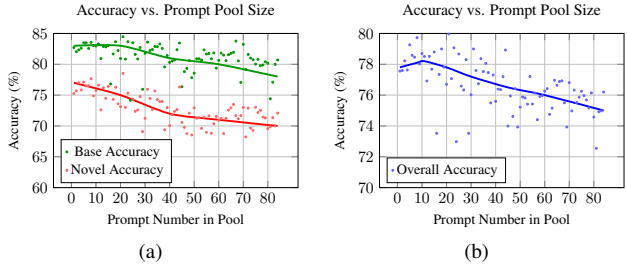
\begin{figure}[ht]
    \centering
    % 在figure环境开始处添加以下命令来设置子图标题的格式
    \captionsetup[subfigure]{justification=centering}  % 这会使子图标题居中
    \begin{subfigure}[t]{0.48\columnwidth}
        \centering
        \begin{tikzpicture}[scale=0.6]  % 使用scale缩放整个图
            \begin{axis}[
                title={Accuracy vs. Prompt Pool Size},
                title style={font=\normalsize},  % 减小标题字体
                xlabel={Prompt Number in Pool},
                ylabel={Accuracy (\%)},
                xlabel style={font=\small},  % 减小轴标签字体
                ylabel style={font=\small},
                grid=both,
                grid style={line width=.1pt, draw=gray!20},
                major grid style={line width=.2pt,draw=gray!50},
                xtick={0,10,20,30,40,50,60,70,80},
                ytick={60,65,70,75,80,85},
                ymin=60, ymax=85,
                width=7cm,  % 减小宽度
                height=5cm,  % 减小高度
                legend pos=south west,
                legend style={
                    font=\small,  % 减小图例字体
                    /tikz/every mark/.append style={scale=1}
                },
                mark options={scale=0.3}
            ]
             % Base performance scatter
            \addplot[green!60!black, only marks, mark=*] coordinates {
            (1,82.716)
            (2,82.053)
            (3,82.332)
            (4,82.612)
            (5,83.485)
            (6,83.485)
            (7,82.577)
            (8,83.554)
            (9,83.031)
            (10,82.961)
            (11,83.310)
            (12,80.936)
            (13,82.891)
            (14,82.821)
            (15,83.031)
            (16,76.816)
            (17,81.669)
            (18,83.659)
            (19,82.228)
            (20,80.936)
            (21,84.462)
            (22,80.831)
            (23,83.659)
            (24,74.162)
            (25,83.485)
            (26,81.599)
            (27,80.866)
            (28,82.716)
            (29,75.943)
            (30,81.774)
            (31,83.589)
            (32,82.228)
            (33,82.402)
            (34,82.891)
            (35,82.856)
            (36,83.310)
            (37,82.577)
            (38,79.574)
            (39,81.425)
            (40,80.307)
            (41,80.517)
            (42,80.691)
            (43,82.367)
            (44,83.450)
            (45,76.362)
            (46,78.806)
            (47,79.330)
            (48,80.587)
            (49,77.828)
            (50,78.561)
            (51,80.901)
            (52,80.901)
            (53,80.691)
            (54,82.332)
            (55,81.669)
            (56,81.739)
            (57,80.901)
            (58,81.844)
            (59,79.365)
            (60,80.622)
            (61,81.809)
            (62,81.494)
            (63,80.587)
            (64,81.564)
            (65,81.774)
            (66,81.948)
            (67,83.764)
            (68,80.552)
            (69,77.060)
            (70,83.240)
            (71,79.155)
            (72,81.599)
            (73,81.006)
            (74,80.726)
            (75,80.028)
            (76,80.796)
            (77,79.784)
            (78,79.888)
            (79,80.412)
            (80,80.168)
            (81,74.406)
            (82,77.095)
            (83,79.993)
            (84,80.691)
            };
            \addlegendentry{Base Accuracy}
            % Base trend line
            \addplot[green!60!black, thick, smooth, line width=1.2pt, forget plot] coordinates {
            (1,83)
            (20,83)
            (40,81)
            (60,80)
            (84,78)
            };
            % \addlegendentry{Base Accuracy}
            
            % Novel performance scatter
            \addplot[red!60, only marks, mark=*] coordinates {
            (1,75.268)
            (2,75.699)
            (3,76.993)
            (4,75.837)
            (5,77.114)
            (6,76.493)
            (7,76.527)
            (8,77.667)
            (9,74.888)
            (10,75.734)
            (11,76.614)
            (12,72.610)
            (13,76.182)
            (14,74.474)
            (15,76.407)
            (16,67.294)
            (17,74.318)
            (18,76.269)
            (19,73.593)
            (20,73.127)
            (21,78.530)
            (22,73.007)
            (23,75.319)
            (24,67.622)
            (25,74.353)
            (26,72.368)
            (27,72.575)
            (28,74.767)
            (29,69.089)
            (30,71.246)
            (31,75.941)
            (32,74.456)
            (33,72.661)
            (34,75.682)
            (35,74.301)
            (36,73.939)
            (37,73.956)
            (38,69.382)
            (39,72.972)
            (40,70.055)
            (41,71.833)
            (42,72.368)
            (43,72.661)
            (44,76.320)
            (45,70.573)
            (46,69.779)
            (47,68.744)
            (48,72.075)
            (49,68.536)
            (50,72.938)
            (51,71.246)
            (52,71.332)
            (53,69.606)
            (54,71.419)
            (55,68.968)
            (56,71.626)
            (57,68.813)
            (58,70.349)
            (59,71.540)
            (60,70.918)
            (61,71.212)
            (62,68.761)
            (63,71.229)
            (64,72.558)
            (65,72.938)
            (66,73.490)
            (67,71.695)
            (68,71.160)
            (69,71.695)
            (70,73.007)
            (71,68.226)
            (72,72.903)
            (73,69.658)
            (74,69.279)
            (75,73.024)
            (76,72.178)
            (77,72.523)
            (78,71.436)
            (79,71.298)
            (80,69.572)
            (81,69.900)
            (82,71.332)
            (83,70.003)
            (84,72.075)
            };
            \addlegendentry{Novel Accuracy}
            % Novel trend line
            \addplot[red, thick, smooth, line width=1.2pt, forget plot] coordinates {
            (1,77)
            (20,75)
            (40,72)
            (60,71)
            (84,70)
            };
            % 这里保持原有的数据点和曲线代码不变
        \end{axis}
    \end{tikzpicture}
    \caption{}
    \label{tab:Base_Novel_vv_PromptLength}
    \end{subfigure}
    \hfill  % 在两个subfigure之间添加弹性间距
    \begin{subfigure}[t]{0.48\columnwidth}
        \centering
        \begin{tikzpicture}[scale=0.6]
            \begin{axis}[
                title={Accuracy vs. Prompt Pool Size},
                title style={font=\normalsize},
                xlabel={Prompt Number in Pool},
                ylabel={Accuracy (\%)},
                xlabel style={font=\small},
                ylabel style={font=\small},
                grid=both,
                grid style={line width=.1pt, draw=gray!20},
                major grid style={line width=.2pt,draw=gray!50},
                xtick={0,10,20,30,40,50,60,70,80},
                ytick={70,72,74,76,78,80},
                ymin=70, ymax=80,
                width=7cm,
                height=5cm,
                legend pos=south west,
                legend style={font=\small},
                mark options={scale=0.3}
            ]
            % Overall performance
            \addplot[blue!60, only marks, mark=*] coordinates {
            (1,77.559)
            (2,77.596)
            (3,78.223)
            (4,77.629)
            (5,79.260)
            (6,78.580)
            (7,78.461)
            (8,79.727)
            (9,78.023)
            (10,78.700)
            (11,78.508)
            (12,76.917)
            (13,78.517)
            (14,77.681)
            (15,78.795)
            (16,73.523)
            (17,77.343)
            (18,78.998)
            (19,77.811)
            (20,76.700)
            (21,79.965)
            (22,77.068)
            (23,79.067)
            (24,72.978)
            (25,78.669)
            (26,76.587)
            (27,76.335)
            (28,78.305)
            (29,73.525)
            (30,76.023)
            (31,78.994)
            (32,77.598)
            (33,76.738)
            (34,78.365)
            (35,77.526)
            (36,78.059)
            (37,77.462)
            (38,75.497)
            (39,77.309)
            (40,75.596)
            (41,76.392)
            (42,76.358)
            (43,77.198)
            (44,78.925)
            (45,73.993)
            (46,75.946)
            (47,74.618)
            (48,76.420)
            (49,73.928)
            (50,75.222)
            (51,75.879)
            (52,75.816)
            (53,75.417)
            (54,77.205)
            (55,75.426)
            (56,76.130)
            (57,74.408)
            (58,75.911)
            (59,75.967)
            (60,76.257)
            (61,76.419)
            (62,74.754)
            (63,75.153)
            (64,76.966)
            (65,76.638)
            (66,76.881)
            (67,76.928)
            (68,75.501)
            (69,73.931)
            (70,76.808)
            (71,74.974)
            (72,76.325)
            (73,75.518)
            (74,75.865)
            (75,76.250)
            (76,75.971)
            (77,75.683)
            (78,75.176)
            (79,76.142)
            (80,74.577)
            (81,72.551)
            (82,74.924)
            (83,74.977)
            (84,76.195)
            };
            % \addlegendentry{Overall Accuracy}
            \addlegendentry{Overall Accuracy}
            % Overall trend line with more natural curve
            \addplot[blue, thick, smooth, line width=1.2pt, forget plot] coordinates {
            (1,77.8)
            (8,78.1)
            (11,78.2)
            (20,77.8)
            (30,77.2)
            (45,76.5)
            (60,76)
            (84,75)
            };
        \end{axis}
    \end{tikzpicture}
    \caption{}
    \label{tab:Overall_vv_PromptLength}
    \end{subfigure}
   \caption{Performance analysis with varying prompt pool size: (a) Base-session (green) and novel-session (red) accuracies; (b) Overall accuracy (blue). The base-session accuracy initially increases with the increase of prompt, while the novel-session accuracy drops sharply, showing the difference caused by the number of training data. It reveals that abundant base-session samples enable robust prompt pool learning, while limited novel-session samples lead to performance degradation with increased prompts.}
    \label{fig:performance_analysis}
\end{figure}

%-------------------------------------------------------------------------
However, contrary to expectations, integrating the prompt pool with VPT leads to a notable degradation in novel-session performance. As shown in Figure~\ref{fig:performance_analysis}(a), while base session accuracy marginally improves with the increase of prompt number in the pool, novel session performance sharply decreases, leading to an overall performance degradation.
Specifically, base classes, which benefit from abundant training data, demonstrate strong resilience to increased prompt numbers, maintaining stable performance up to 30 prompts before showing a gradual decline. In contrast, novel classes, with their limited training samples, exhibit immediate and consistent performance deterioration as more prompts are added. The overall accuracy (Figure~\ref{fig:performance_analysis}(b)) reaches its peak at 8-10 selected prompts before declining, suggesting an optimal balance point in prompt selection. 

The difference between the base and novel sessions inspires us to hypothesize that 
%the model's capacity to effectively utilize prompts is primarily determined by the size of training data between the base and novel sessions
\textbf{the model's best number of prompts is primarily determined by the size of training data}, and the different amounts of base- and novel-session training data make the performance decrease at different numbers of prompts.
To further understand it, we then examine how varying numbers of prompts influence the model's learning.
%-------------------------------------------------------------------------
%-------------------------------------------------------------------------

\vspace{-5pt}
\subsection{Limited Training Samples Lead to Token-Dimension Conflict and Saturation}

%-------------------------------------------------------------------------
\begin{figure}[t]
    \centering
    % Base Session (a)
    \begin{minipage}{\columnwidth}
        \centering
        % 第一行图片
        \begin{minipage}{\columnwidth}
            \centering
            \includegraphics[width=0.13\columnwidth]{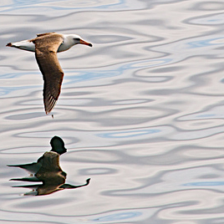}
            \includegraphics[width=0.13\columnwidth]{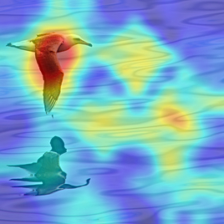}
            \includegraphics[width=0.13\columnwidth]{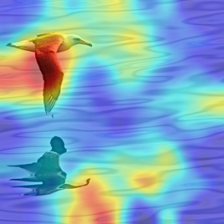}
            \includegraphics[width=0.13\columnwidth]{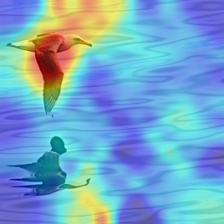}
            \includegraphics[width=0.13\columnwidth]{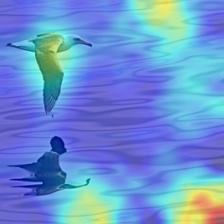}
        \end{minipage}
        
        \vspace{0.3mm}
        % 第二行图片（带标识base）
        \begin{minipage}{\columnwidth}
            \centering
            \makebox[0mm][r]{\raisebox{0.5\height}{\makebox[12mm][l]{Base}}}%
            \includegraphics[width=0.13\columnwidth]{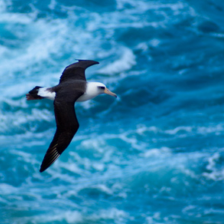}
            \includegraphics[width=0.13\columnwidth]{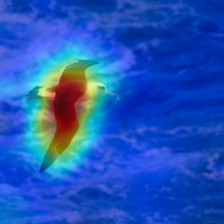}
            \includegraphics[width=0.13\columnwidth]{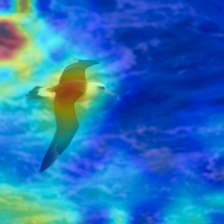}
            \includegraphics[width=0.13\columnwidth]{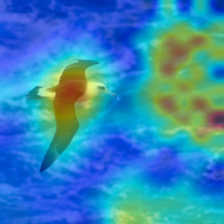}
            \includegraphics[width=0.13\columnwidth]{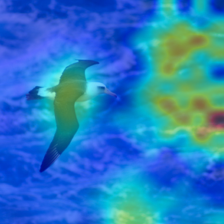}
        \end{minipage}
        
        \vspace{0.3mm}
        % 第三行图片
        \begin{minipage}{\columnwidth}
            \centering
            \includegraphics[width=0.13\columnwidth]{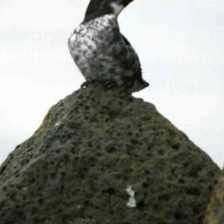}
            \includegraphics[width=0.13\columnwidth]{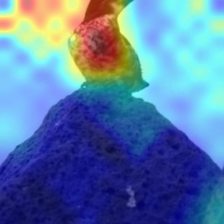}
            \includegraphics[width=0.13\columnwidth]{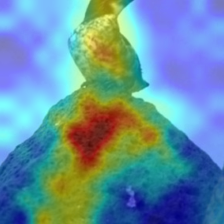}
            \includegraphics[width=0.13\columnwidth]{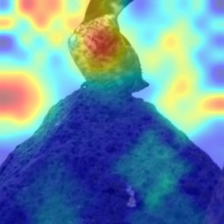}
            \includegraphics[width=0.13\columnwidth]{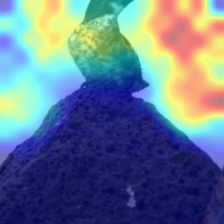}
        \end{minipage}
    \end{minipage}
    
    \vspace{0.3mm}
    \hrulefill
    \vspace{0.3mm}
    
    % Novel Session (b)
    \begin{minipage}{\columnwidth}
        \centering
        \begin{minipage}{\columnwidth}
            \centering
            \includegraphics[width=0.13\columnwidth]{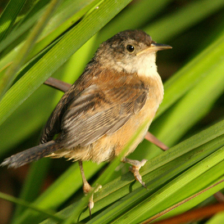} 
            \includegraphics[width=0.13\columnwidth]{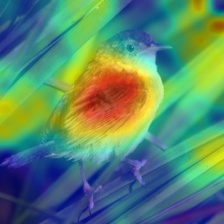} 
            \includegraphics[width=0.13\columnwidth]{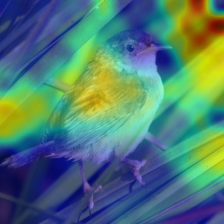} 
            \includegraphics[width=0.13\columnwidth]{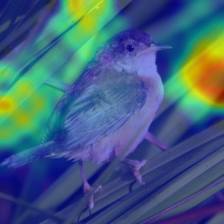} 
            \includegraphics[width=0.13\columnwidth]{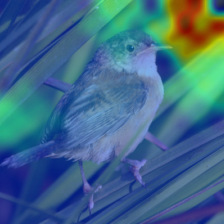}
        \end{minipage}

        \vspace{0.3mm}
        \begin{minipage}{\columnwidth}
            \centering
            \makebox[0mm][r]{\raisebox{0.5\height}{\makebox[12mm][l]{Novel}}}%
            \includegraphics[width=0.13\columnwidth]{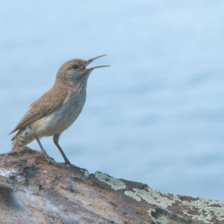} 
            \includegraphics[width=0.13\columnwidth]{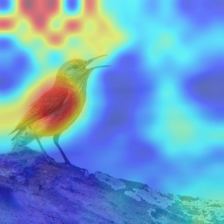} 
            \includegraphics[width=0.13\columnwidth]{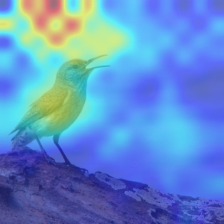} 
            \includegraphics[width=0.13\columnwidth]{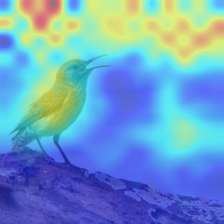} 
            \includegraphics[width=0.13\columnwidth]{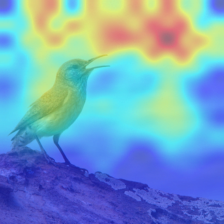}
        \end{minipage}

        \vspace{0.3mm}
        \begin{minipage}{\columnwidth}
            \centering
            \includegraphics[width=0.13\columnwidth]{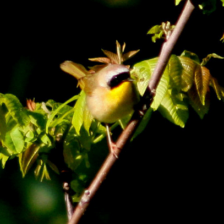} 
            \includegraphics[width=0.13\columnwidth]{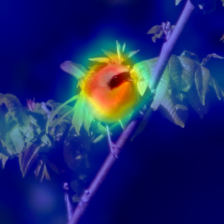} 
            \includegraphics[width=0.13\columnwidth]{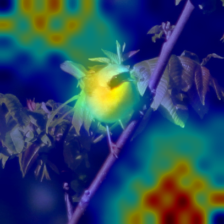} 
            \includegraphics[width=0.13\columnwidth]{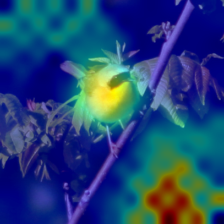} 
            \includegraphics[width=0.13\columnwidth]{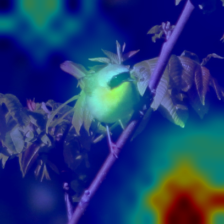}
        \end{minipage}
    \end{minipage}
    
    \vspace{1.2mm}
    % 只在最后写一次标签
    {\footnotesize
    \makebox[0.13\columnwidth]{$original$}
    \makebox[0.13\columnwidth]{$p=1$}
    \makebox[0.13\columnwidth]{$p=3$}
    \makebox[0.13\columnwidth]{$p=5$}
    \makebox[0.13\columnwidth]{$p=7$}
    }\vspace{0.1cm}
    \caption{Class token attention maps with increasing number of selected prompts ($p$) from pool. 
    %Base Session results show gradual attention drift, while Novel Session results show early attention dispersion, revealing how excessive prompts cause attention drift from discriminative features to background elements.
    As the prompt number increases, the base-session model initially focuses on discriminative regions of the images, and then drifts the attention to task-irrelevant regions. In contrast, the novel-session model focuses on the task-irrelevant regions with even a small number of prompts, showing the saturation caused by limited training samples.
    }\vspace{0.2cm}
    \label{fig:ClassToken_vv_Promptlength}
\end{figure}

%-------------------------------------------------------------------------
To investigate these mechanisms, we study the attention map of the CLS token for the global attention of the model, and the attention map of each prompt in the pool for the individual attention of each prompt, 
particularly focusing on the asymmetric effects between base and novel classes.

\subsubsection{Global Attention of the CLS token} 
As shown in Figure~\ref{fig:ClassToken_vv_Promptlength}, for base-session data, when introducing the first few prompts (1-2) in the pool, the model demonstrates enhanced focus on discriminative foreground features, leading to improved classification performance. However, as more prompts are selected (3-5), the attention mechanism begins to disperse between foreground and background regions. This dispersion becomes more pronounced with extensive prompt selection ($\geq$ 9), where attention predominantly shifts to background elements while crucial foreground features receive minimal focus.  

For novel-session data, however, the attention drift phenomenon emerges significantly earlier. Even with minimal prompt selection (p=1), the model's attention already disperses between foreground and background regions. As more prompts are incorporated (3-7), the attention becomes increasingly scattered across background elements, which can be attributed to the limited sample size of novel classes.

This result reveals that excessive prompt selection causes attention drift from discriminative features to background elements, which means \textbf{the effective information that the prompt pool contains is limited by the training data size, i.e., the prompt pool saturates}, leading to the observed performance degradation in Figure~\ref{fig:performance_analysis}.

\subsubsection{Individual Attention of Each Prompt in the Pool} 
\begin{figure}[t]
    \centering
    \begin{minipage}{\columnwidth}
        \centering
        \makebox[0mm][r]{\raisebox{0.5\height}{(a)}\hspace{2mm}}%
        \includegraphics[width=0.13\columnwidth]{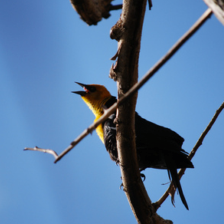}
        \includegraphics[width=0.13\columnwidth]{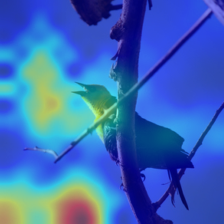}
        \includegraphics[width=0.13\columnwidth]{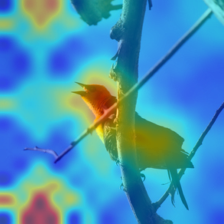}
        \includegraphics[width=0.13\columnwidth]{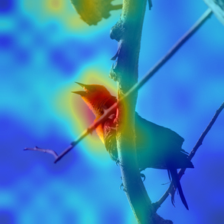}
        \includegraphics[width=0.13\columnwidth]{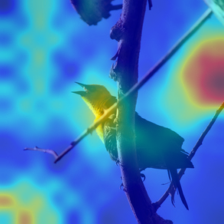}
        \includegraphics[width=0.13\columnwidth]{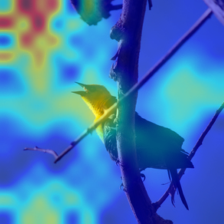}
    \end{minipage}

    \vspace{0.4mm}
    \hrulefill
    \vspace{0.4mm}
    
    \begin{minipage}{\columnwidth}
        \centering
        \makebox[0mm][r]{\raisebox{0.5\height}{(b)}\hspace{2mm}}%
        \includegraphics[width=0.13\columnwidth]{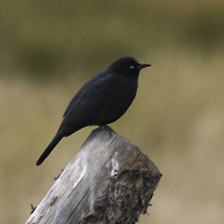} 
        \includegraphics[width=0.13\columnwidth]{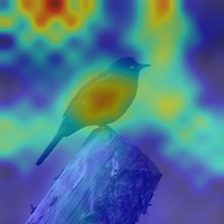} 
        \includegraphics[width=0.13\columnwidth]{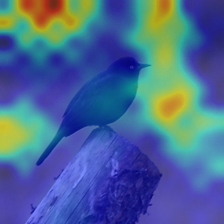} 
        \includegraphics[width=0.13\columnwidth]{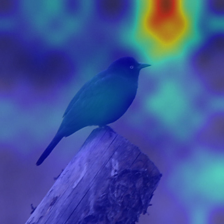} 
        \includegraphics[width=0.13\columnwidth]{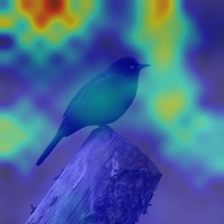}
        \includegraphics[width=0.13\columnwidth]{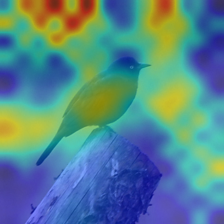}
    \end{minipage}
    
    \vspace{1.5mm}
    {\footnotesize
    \makebox[0.13\columnwidth]{$original$}
    \makebox[0.13\columnwidth]{$1st$}
    \makebox[0.13\columnwidth]{$2nd$}
    \makebox[0.13\columnwidth]{$3rd$}
    \makebox[0.13\columnwidth]{$4th$}
    \makebox[0.13\columnwidth]{$5th$}
    }\vspace{0.1cm}
    \caption{Attention heat maps of prompts selected from pool. The base-session results (a) show attention on both task-relevant and irrelevant regions, while the novel-session results (b) show predominantly background attention, revealing prompts encode task-irrelevant information when the task-relevant ones are saturated.}\vspace{0.1cm}
    \label{fig:prompt_vv_num}
\end{figure}

To better understand how this saturation affects each prompt in the pool, we further study the attention map of each individual prompt.
As in Figure~\ref{fig:prompt_vv_num}, on base-session data, we discover that different prompts in the pool exhibit distinct and sometimes competing attention patterns. Some prompts (e.g., Prompt 3) demonstrate a strong focus on class-discriminative features such as distinctive bird markings, effectively supporting the model's classification task. In contrast, others (e.g., Prompts 1 and 5) show stronger attention to background elements. 
For the novel-session data, we observe a concerning pattern where most prompts (e.g., Prompts 2, 4 and 5) predominantly attend to background regions rather than class-discriminative features, suggesting a failure to capture meaningful object patterns. This phenomenon reveals that the prompts have lost their ability to identify task-relevant features when processing novel categories.
Given the saturation of information in the prompt pool, \textbf{we interpret that each prompt in the pool competes with each other to encode the task-relevant information}, while the loser can only encode those task-irrelevant ones, leading to the overfitting to the training data.
With only scarce training data, there may be no winner, as the available information cannot even fill up a single prompt, leading to the attention on background in Figure~\ref{fig:prompt_vv_num}(b).

%-------------------------------------------------------------------------

%-------------------------------------------------------------------------

\subsection{Conclusion and Discussion} 
Our analysis reveals a fundamental challenge in prompt-based learning: \textbf{given limited training samples, the information that the prompt pool (as appended tokens) encodes could be saturated}.
Therefore, a model cannot utilize infinite prompts to encode the available information. 
Specifically, this saturation leads to a competitive behavior between prompts for encoding the task-relevant information, while losers can only encode task-irrelevant ones, ultimately degrading the model's attention to the input sample. 
Such a problem is exacerbated by the scarce training data on novel sessions, which makes the prompt pool can only encode task-irrelevant information, downgrading novel-session performance.
 
Inspired by this interpretation, given the promising performance of VPT-based method and pool-based prompt learning method, we question that we may not need to add the prompt pool in the token dimension that is saturated, but to take other dimensions for prompting.
Building upon this insight and recent advances in spatial prompting~\cite{huang2023diversityawaremetavisualprompting, liu2024insvp, kirillov2023segment}, we propose to leverage the spatial domain as an alternative dimension for the pool, leading to our LGSP-Prompt framework.\vspace{-3pt}

%-------------------------------------------------------------------------
%-------------------------------------------------------------------------

\begin{figure*}[t]
    \centering
    \includegraphics[width=0.97\textwidth]{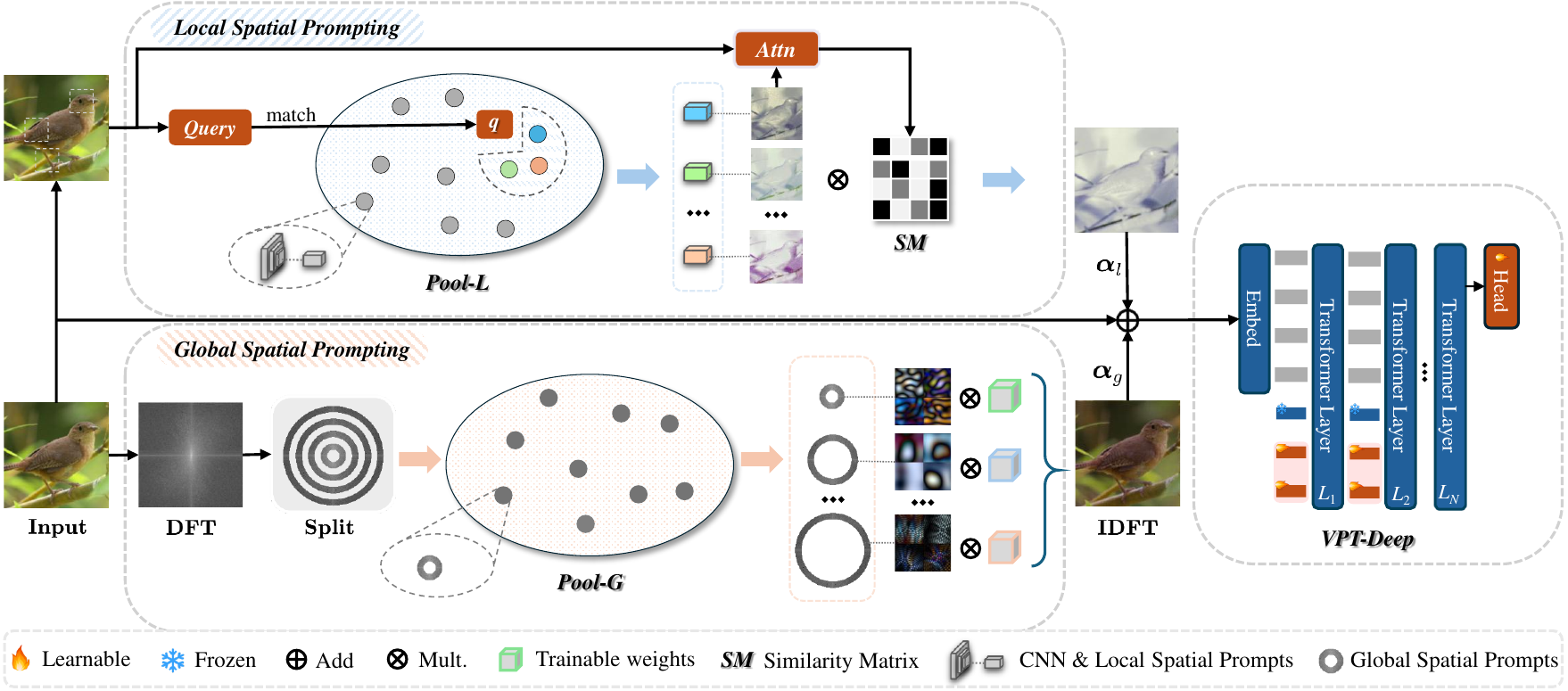}
    \caption{Overview of LGSP-Prompt framework. Our approach leverages both local and global spatial information through three key components: (1) Local Spatial Prompting with dynamic prompt pool for fine-grained feature extraction, (2) Global Spatial Prompting operating in frequency domain for holistic representation, and (3) VPT-Deep module for efficient knowledge preservation. These components are synergistically combined through adaptive weighting mechanisms ($\alpha_{l}$, $\alpha_{g}$) and residual connections to enhance FSCIL performance.}
    \label{fig:overview}
    \vspace{-10pt}
\end{figure*}

\vspace{0.1cm}
\section{Methodology}
\label{sec:method}
Our LGSP-Prompt framework addresses token-dimension saturation through dual-perspective spatial adaptation: local spatial prompting for fine-grained features and global spatial prompting for holistic representations, integrated by VPT-Deep. As illustrated in Figure~\ref{fig:overview}, these components work synergistically to enable effective feature learning while avoiding token-dimension conflicts.
%-------------------------------------------------------------------------
\subsection{Local Spatial Prompting}
\label{sec:Local}

\subsubsection{Prompt Pool Construction}
To effectively capture fine-grained spatial features while preventing catastrophic forgetting, we design a local spatial prompting mechanism that maintains a pool of diverse spatial prompts $\mathcal{P}$. Each element in the pool consists of convolutional filters and their corresponding local spatial prompts, enabling the capture of position-sensitive characteristics crucial for visual recognition.

Given an input image \(X \in \mathbb{R}^{B \times C \times H \times W}\), where \(B\) is the batch size, \(C=3\) represents the RGB channels, and \(H, W\) are spatial dimensions, we construct a pool of \(M\) prompt generators and their corresponding prompts:
\begin{equation}
P_i = f_i(X) \in \mathbb{R}^{d_p}, \quad i \in \{1, 2, \dots, M\}
\label{eq:prompt_gen}
\end{equation}
where each $f_i$ is implemented as a two-layer convolutional network with dropout for robust feature generation.
\subsubsection{Dynamic Selection and Integration}
To adaptively utilize the most relevant spatial features, we design a two-stage selection process that dynamically identifies and combines the most pertinent spatial prompts for each input. First, we obtain the query key through class token extraction and feature projection:
\begin{equation}
\mathbf{q} = g(f(X)[0,:])
\label{eq:query}
\end{equation}
where $f(\cdot)$ denotes the Vision Transformer backbone, $[0,:]$ selects the class token, and $g(\cdot)$ is a feature projection layer.

The dynamic selection of the most relevant prompts is to select $n_s$ prompts from the pool using cosine similarity as the matching criterion:
\begin{equation}
\mathcal{T} = \underset{\{s_i\}_{i=1}^{n_s} \subseteq [1,M]}{\text{argmax}} \sum_{i=1}^{n_s} \gamma(\mathbf{q}, P_{s_i})
\label{eq:topk_selection}
\end{equation}
where $\gamma(\cdot,\cdot)$ denotes the cosine similarity between query feature and prompt features.
The selected prompts are weighted by an attention mechanism with parameter $\tau$:
\begin{equation}
w_i = \frac{\exp(\gamma(k, P_i)/\tau)}{\sum_{j \in \mathcal{T}} \exp(\gamma(k, P_j)/\tau)},\,i \in \mathcal{T}
\label{eq:weights_norm}
\end{equation}

Finally, we aggregate the weighted prompts to obtain a comprehensive local spatial representation:
\begin{equation}
P_{\text{local}} = \sum_{i \in \mathcal{T}} w_i \odot P_i
\label{eq:aggregation_final}
\end{equation}
where $\odot$ denotes element-wise multiplication. This adaptive mechanism allows model to dynamically focus on core spatial features while maintaining past knowledge.
%-------------------------------------------------------------------------
%-------------------------------------------------------------------------
\subsection{Global Spatial Prompting}
\label{sec:Global}
Complementing the local spatial features, we propose a global spatial prompting mechanism that operates in the frequency domain, enabling holistic pattern learning while avoiding token-dimension saturation.

\subsubsection{Frequency Domain Transformation}
Given an input image \( X \in \mathbb{R}^{B \times C \times H \times W} \), we first apply Discrete Fourier Transform (DFT) to obtain its frequency representation and center the spectrum, denoted as \(\mathcal{F}_c(X)\). The centered spectrum naturally organizes frequency components from low to high frequencies, center to periphery.

\subsubsection{Concentric Ring Partition}
To systematically process different frequency components, we partition the frequency spectrum into $K$ concentric rings. We first construct a distance matrix $D$ that measures the distance from each point to the spectrum center:
\begin{equation}
D_{y,x} = \sqrt{(y - \frac{H}{2})^2 + (x - \frac{W}{2})^2}
\end{equation}
where \((y,x)\) denotes spatial coordinates. Based on this distance matrix, we define $K$ rings with radii:
\begin{equation}
r_k = \frac{k}{K}R_{\text{max}}, \quad k \in \{1, 2, \dots, K\}
\end{equation}
where \(R_{\text{max}} = \sqrt{(H/2)^2 + (W/2)^2}\) represents the maximum radius from center (low frequency) to corner (high frequency). This structure creates a series of rings that form the spatial prompt pool in the frequency domain.

\subsubsection{Differentiable Ring Mask Generation}
For each ring $k$, we obtain a differentiable mask by:
\begin{equation}
M_k = \sigma(-\beta(D - r_k)) - \sigma(-\beta(D - r_{k-1}))
\end{equation}
where \(\sigma(x) = \frac{1}{1 + e^{-x}}\) is the sigmoid function ensuring smooth transitions, \(\beta\) controls the sharpness of ring boundaries, and \(r_{k-1}\) is the inner radius of the $k$-th ring (\(r_0 = 0\)). This formulation creates a series of overlapping ring-shaped masks, where each mask smoothly transitions from 1 within its designated ring to 0 outside.

\subsubsection{Adaptive Frequency Enhancement}
These frequency-based prompts are weighted through learnable parameters $w_k \sim \mathcal{N}(0, 1)$ with temperature $\tau$. The enhanced frequency representation is obtained by:
\begin{equation}
\mathcal{F}_{\text{enhanced}}(X) = \mathcal{F}_c(X) \odot \sum_{k=1}^{K} \text{softmax}(w_k \tau) M_k
\end{equation}
where \(\odot\) denotes element-wise multiplication, enabling adaptive emphasis of different frequency bands.
%-------------------------------------------------------------------------
\subsection{Adaptive Feature Integration}
\label{sec:Fusion}
To effectively combine the complementary strengths of local spatial features $P_{\text{local}}$ and global frequency features $X_{\text{global}}$, we employ an adaptive integration mechanism:
\begin{equation}
X_{\text{final}} = X + \alpha_l P_{\text{local}} + \alpha_g X_{\text{global}},
\end{equation}
where $\alpha_l$ and $\alpha_g$ are learnable parameters that balance the dual-perspective spatial adaptation without introducing token-dimension conflicts. This integration strategy is maintained during both training and inference, preserving the complementary benefits of all prompts. Following FSCIL practice, we finetune the classifier and spatial prompting with a small learning rate in novel sessions.\vspace{-2pt}
\begin{table}[t]
    \centering
    \small
    \setlength{\tabcolsep}{3pt}
    \begin{tabular}{l|l|r|c|c}
        \hline
        Dataset & Domain & Images & Base & Novel \\
        \hline
        CUB-200 & Birds & 11,788 & 100w-30s & 10×(10w-5s) \\
        iNF200 & Fungi & 10,000 & 100w-50s & 10×(10w-5s) \\
        FGVCAircraft & Aircraft & 10,000 & 50w-30s & 10×(5w-5s) \\
        \hline
    \end{tabular}
    \caption{Dataset configurations for base and novel (10 sessions) where N and K denote categories and samples in N-way-K-shot.}
    
    \label{tab:dataset_setup}
    \vspace{-5pt}
\end{table}

\section{Experiments}
\label{sec:exp}

\begin{table*}[t]
  \centering
  \footnotesize
  \setlength{\tabcolsep}{2pt}
  \begin{tabular}{@{}l c*{11}{c}c@{}}
    \toprule
    Method & Venue & S0 & S1 & S2 & S3 & S4 & S5 & S6 & S7 & S8 & S9 & S10 & Avg. $\uparrow$ \\
    \midrule
    Fine-Tuning + Proto & N/A & 84.21 & 66.43 & 25.00 & 25.44 & 16.19 & 4.58 & 1.42 & 1.49 & 3.62 & 5.50 & 3.79 & 21.60 \\
    iCaRL~\cite{rebuffi2017icarlincrementalclassifierrepresentation} & CVPR'17 & 82.43 & 79.32 & 68.74 & 59.93 & 61.62 & 59.91 & 57.83 & 57.34 & 55.42 & 52.73 & 55.92 & 63.83 \\
    RDI~\cite{zhou2024delvebasenovelconfusionredundancy} & IJCAI'24 & 80.13 & 76.55 & 73.21 & 69.37 & 67.83 & 65.74 & 64.91 & 63.37 & 61.43 & 61.41 & 60.20 & 67.65 \\
    CEC~\cite{Zhang_2021_CVPR} & CVPR'21 & 81.82 & 79.53 & 78.42 & 75.54 & 76.31 & 74.83 & 74.41 & 74.62 & 74.23 & 73.91 & 73.84 & 76.13 \\
    FOSTER~\cite{wang2022fosterfeatureboostingcompression} & ECCV'22 & 85.02 & 83.43 & 77.41 & 71.52 & 69.93 & 66.34 & 65.52 & 63.07 & 62.92 & 62.03 & 60.42 & 69.78 \\
    FACT~\cite{zhou2022forward} & CVPR'22 & 84.32 & 81.23 & 79.14 & 75.13 & 75.42 & 73.31 & 72.43 & 72.52 & 71.41 & 71.12 & 70.91 & 75.18 \\
    CLOM~\cite{zou2022marginbasedfewshotclassincrementallearning} & NIPS'22 & 83.28 & 81.85 & 79.61 & 77.79 & 76.34 & 74.64 & 73.62 & 72.82 & 71.24 & 71.33 & 70.50 & 75.73 \\
    NC-FSCIL~\cite{yang2023neural} & ICLR'23 & 83.52 & 80.92 & 80.14 & 77.83 & 77.81 & 76.96 & 76.72 & 74.78 & 74.18 & 73.92 & 73.80 & 77.32 \\
    WaRP~\cite{kim2023warping} & ICLR'23 & 82.74 & 80.21 & 79.06 & 77.80 & 77.78 & 76.81 & 76.82 & 74.61 & 74.13 & 74.02 & 73.36 & 77.03 \\
    TEEN~\cite{wang2023fewshotclassincrementallearningtrainingfree} & NIPS'23 & 84.03 & 81.52 & 80.91 & 78.34 & 78.32 & 77.24 & 77.13 & 75.42 & 75.51 & 75.13 & 75.61 & 78.11 \\
    % RDI~\cite{zhou2024delvebasenovelconfusionredundancy} & IJCAI'24 & 82.31 & 80.65 & 79.85 & 76.96 & 77.08 & 74.85 & 74.99 & 75.40 & 74.29 & 74.69 & 74.38 & 76.86 \\
    Comp-FSCIL~\cite{zou2024compositionalfewshotclassincrementallearning} & ICML'24 & 83.67 & 81.73 & 79.03 & 78.04 & 77.73 & 75.52 & 74.32 & 74.55 & 73.35 & 73.15 & 72.80 & 76.72 \\
    Yourself~\cite{tang2024rethinking} & ECCV'24 & 82.31 & 80.65 & 79.85 & 76.96 & 77.08 & 74.85 & 74.99 & 75.40 & 74.29 & 74.69 & 74.38 & 76.86 \\
    PriViLege~\cite{park2024pretrainedvisionlanguagetransformers} & CVPR'24 & 82.21 & 81.25 & 80.45 & 77.76 & 77.78 & 75.95 & 75.69 & 76.00 & 75.19 & 75.19 & 75.08 & 77.50 \\
    \midrule
    L2P\dag~\cite{wang2022learning} & CVPR'22 & 82.47 & 81.23 & 79.01 & 76.89 & 76.21 & 74.73 & 74.19 & 74.11 & 72.73 & 73.02 & 73.67 & 76.21 \\
    DualPrompt\dag~\cite{wang2022dualprompt} & ECCV'22 & 83.51 & 82.27 & 80.93 & 79.57 & 78.63 & 77.09 & 76.31 & 77.03 & 75.79 & 76.17 & 76.53 & 78.53 \\
    CODA-Prompt\dag~\cite{wang2023codaprompt} & CVPR'23 & 79.61 & 78.12 & 76.42 & 75.68 & 75.02 & 73.19 & 72.58 & 72.81 & 72.07 & 72.49 & 72.97 & 74.63 \\
    \midrule
    \textbf{LGSP-Prompt} & Ours & \textbf{85.72} & \textbf{84.31} & \textbf{83.21} & \textbf{81.33} & \textbf{81.80} & \textbf{80.33} & \textbf{79.89} & \textbf{80.09} & \textbf{79.18} & \textbf{79.74} & \textbf{79.72} & \textbf{81.39} \\
    \bottomrule
  \end{tabular}
  \caption{\small Accuracy on CUB-200 benchmark. \textbf{Avg:} average of all sessions. \textbf{Bold:} the best performance. \dag: prompt pool-based methods.}
  \label{tab:cub200_results}
    \vspace{-2pt}
\end{table*}

\begin{table*}[t]
  \centering
  \footnotesize
  \setlength{\tabcolsep}{2.5pt}
  \begin{tabular}{@{}l c*{11}{c}c@{}}
    \toprule
    Method & Venue & S0 & S1 & S2 & S3 & S4 & S5 & S6 & S7 & S8 & S9 & S10 & Avg. $\uparrow$ \\
    \midrule
    Yourself~\cite{tang2024rethinking} & ECCV'24 & 26.29 & 20.37 & 17.05 & 14.82 & 13.16 & 11.87 & 10.83 & 9.97 & 9.24 & 8.62 & 8.08 & 13.66 \\
    Approximation~\cite{Wang2024OnTA} & ECCV'24 & 21.67 & 19.48 & 18.90 & 17.57 & 16.50 & 16.68 & 15.82 & 15.21 & 14.20 & 13.85 & 12.97 & 16.62 \\
    PriViLege~\cite{park2024pretrainedvisionlanguagetransformers} & CVPR'24 & 22.75 & 20.79 & 21.85 & 20.96 & 20.27 & 22.32 & 21.53 & 21.67 & 20.43 & 20.18 & 20.04 & 21.16 \\
    \midrule
    \textbf{LGSP-Prompt} & Ours & \textbf{27.67} & \textbf{24.82} & \textbf{26.30} & \textbf{27.38} & \textbf{25.25} & \textbf{26.76} & \textbf{26.37} & \textbf{25.49} & \textbf{25.50} & \textbf{24.83} & \textbf{23.87} & \textbf{25.82} \\
    \bottomrule
  \end{tabular}
  \caption{\small Accuracy on FGVCAircraft benchmark. \textbf{Avg:} average of all sessions. \textbf{Bold:} the best performance.}
  \label{tab:fgvcaircraft_results}
  \vspace{-5pt}
\end{table*}

\begin{table*}[t]
  \centering
  \footnotesize
  \setlength{\tabcolsep}{2.5pt}
  \begin{tabular}{@{}l c*{11}{c}c@{}}
    \toprule
    Method & Venue & S0 & S1 & S2 & S3 & S4 & S5 & S6 & S7 & S8 & S9 & S10 & Avg. $\uparrow$ \\
    \midrule
    Yourself~\cite{tang2024rethinking} & ECCV'24 & 61.90 & 55.73 & 50.50 & 45.31 & 40.93 & 35.07 & 30.75 & 26.18 & 22.72 & 19.47 & 16.10 & 36.97 \\
    Approximation~\cite{Wang2024OnTA} & ECCV'24 & 66.30 & 60.18 & 57.25 & 53.62 & 50.43 & 47.67 & 44.88 & 42.41 & 40.11 & 38.32 & 36.75 & 48.90 \\
    PriViLege~\cite{park2024pretrainedvisionlanguagetransformers} & CVPR'24 & 63.10 & 60.73 & 59.50 & 58.31 & 57.93 & 56.07 & 53.75 & 52.18 & 51.72 & 49.47 & 48.10 & 55.53 \\
    \midrule
    \textbf{LGSP-Prompt} & Ours & \textbf{67.00} & \textbf{63.27} & \textbf{62.83} & \textbf{60.92} & \textbf{59.57} & \textbf{57.67} & \textbf{56.19} & \textbf{53.88} & \textbf{53.83} & \textbf{50.05} & \textbf{48.60} & \textbf{57.62} \\
    \bottomrule
  \end{tabular}
  \caption{\small Accuracy on iNF200 benchmark. \textbf{Avg.} represents the average across all the sessions. \textbf{Bold} values indicate the best performance.}
  \label{tab:inf200_results}
  \vspace{-8pt}
\end{table*}

%-------------------------------------------------------------------------
\subsection{Empirical Setup}
Given that our approach leverages ImageNet as the pre-training dataset, due to its overlapping between CIFAR~\cite{Krizhevsky2009LearningML} and miniImageNet~\cite{vinyals2017matchingnetworksshotlearning}, we evaluate on three fine-grained datasets with distinct visual distributions: CUB-200~\cite{Wah2011TheCB}, iNF200~\cite{VanHorn2018iNaturalist}, and FGVCAircraft~\cite{maji2013finegrainedvisualclassificationaircraft}. We implement our framework on a Vision Transformer backbone pre-trained on ImageNet-21K with VPT-deep configuration~\cite{jia2022visualprompttuning} following~\cite{park2024pretrainedvisionlanguagetransformers}, and compare it against various FSCIL approaches across base and novel sessions, with detailed settings shown in Table~\ref{tab:dataset_setup} and comprehensive implementation details provided in the supplementary material.

%-------------------------------------------------------------------------
%-------------------------------------------------------------------------
\begin{table}[t]
\footnotesize
\setlength{\tabcolsep}{1.8pt}  % 将列间距从 2.5pt 减小到 1.8pt
\centering
\begin{tabular}{@{}l ccc ccc ccc@{}}
\toprule
\multirow{2}{*}{Methods} & \multicolumn{3}{c}{CUB200} & \multicolumn{3}{c}{FGVC} & \multicolumn{3}{c}{iNF200} \\
\cmidrule(lr){2-4} \cmidrule(lr){5-7} \cmidrule(lr){8-10}
& O & B & N & O & B & N & O & B & N \\
\midrule
Vanilla ViT & 61.66 & 68.26 & 61.00 & 15.93 & 16.93 & 15.84 & 30.51 & 35.40 & 30.02 \\
VPT & 79.26 & 83.49 & 78.84 & 20.41 & 24.91 & 19.96 & 51.48 & 60.60 & 50.68 \\
VPT+LSP* & 79.49 & 84.01 & 79.30 & 20.85 & 25.62 & 20.35 & 52.94 & 62.15 & 51.89 \\
VPT+LSP & 80.16 & 84.64 & 79.71 & 21.58 & 28.27 & 20.92 & 56.87 & 67.40 & 55.82 \\
VPT+GSP & 79.89 & 83.59 & 79.52 & 22.18 & 27.43 & 21.66 & 56.51 & 65.60 & 55.60 \\
\midrule
Ours & \textbf{81.39} & \textbf{85.72} & \textbf{80.96} & \textbf{25.82} & \textbf{27.67} & \textbf{25.64} & \textbf{57.62} & \textbf{67.00} & \textbf{56.68} \\
\bottomrule
\end{tabular}
\caption{\small Ablation study on three datasets. LSP/GSP: Local/Global Spatial Prompting, LSP*: single prompt variant. O denotes the overall average accuracy across all sessions, B represents the accuracy on base classes in the first session, and N indicates the average accuracy on novel classes across all subsequent sessions.}
\label{tab:ablation}
\vspace{-8pt}
\end{table}

%-------------------------------------------------------------------------
\subsection{Benchmark Comparisons}
We evaluate our approach against state-of-the-art methods on three fine-grained datasets. As shown in Tables~\ref{tab:cub200_results},~\ref{tab:fgvcaircraft_results}, and~\ref{tab:inf200_results}, our method consistently achieves superior performance. Among \textbf{traditional FSCIL methods}, iCaRL and CEC exhibit severe catastrophic forgetting on CUB-200 (55.92\%), while transformer-based approaches like PriViLege maintain accuracies above 74\%. In the \textbf{prompt pool-based methods}, DualPrompt and CODA-Prompt show promising results but face token-dimension saturation. Our spatial-aware prompt selection achieves 85.72\% initial accuracy and leads on FGVCAircraft and iNF200 (25.82\% and 57.62\%), demonstrating effective capture of both local and global spatial information for fine-grained recognition.\vspace{-2pt}

%-------------------------------------------------------------------------
%-------------------------------------------------------------------------
% 权重
\begin{figure}[t]
    \centering
    \begin{subfigure}[b]{0.18\columnwidth}
        \centering
        \includegraphics[width=\textwidth]{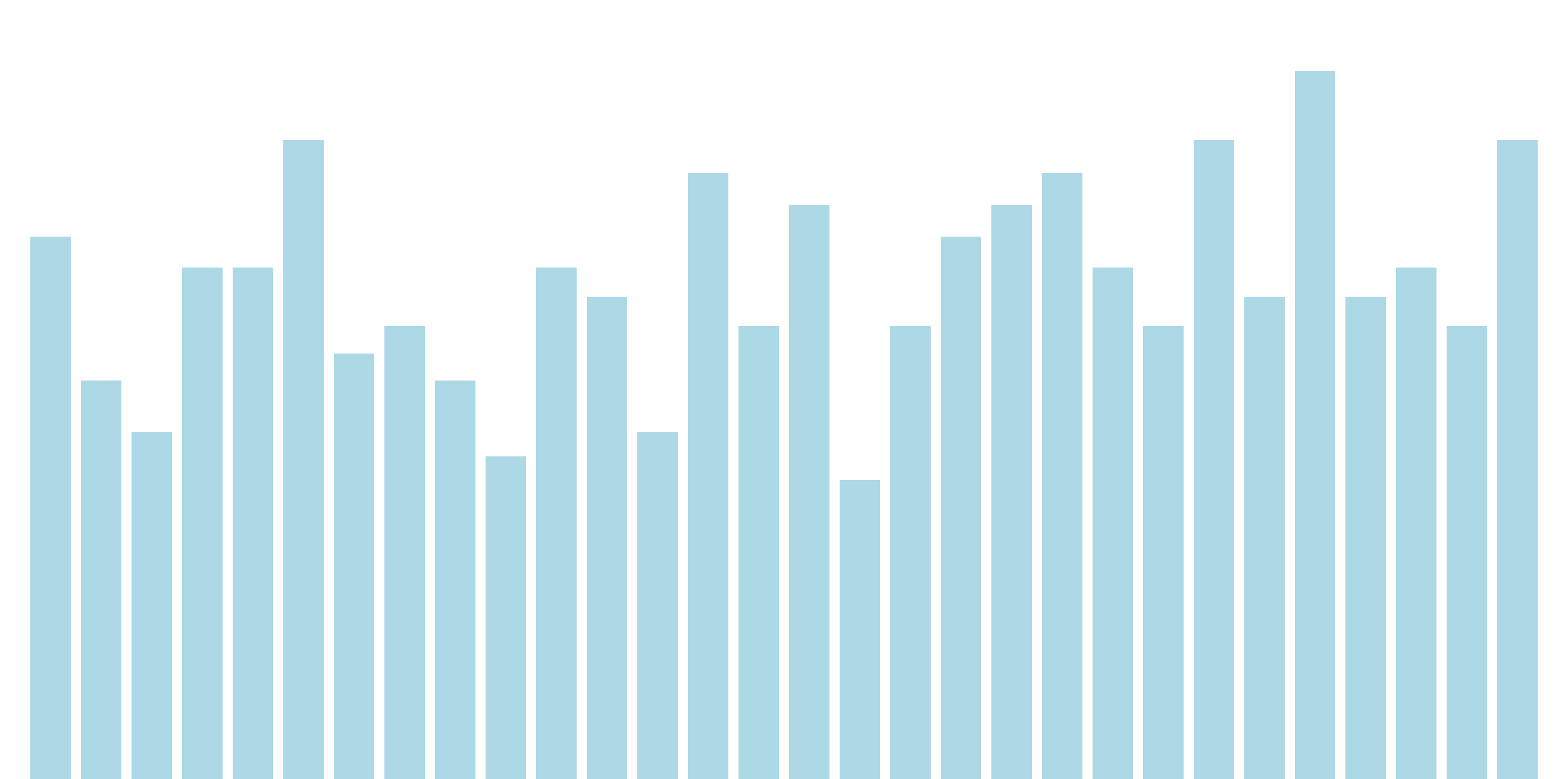}
        \caption{}
    \end{subfigure}
    \hfill
    \begin{subfigure}[b]{0.18\columnwidth}
        \centering
        \includegraphics[width=\textwidth]{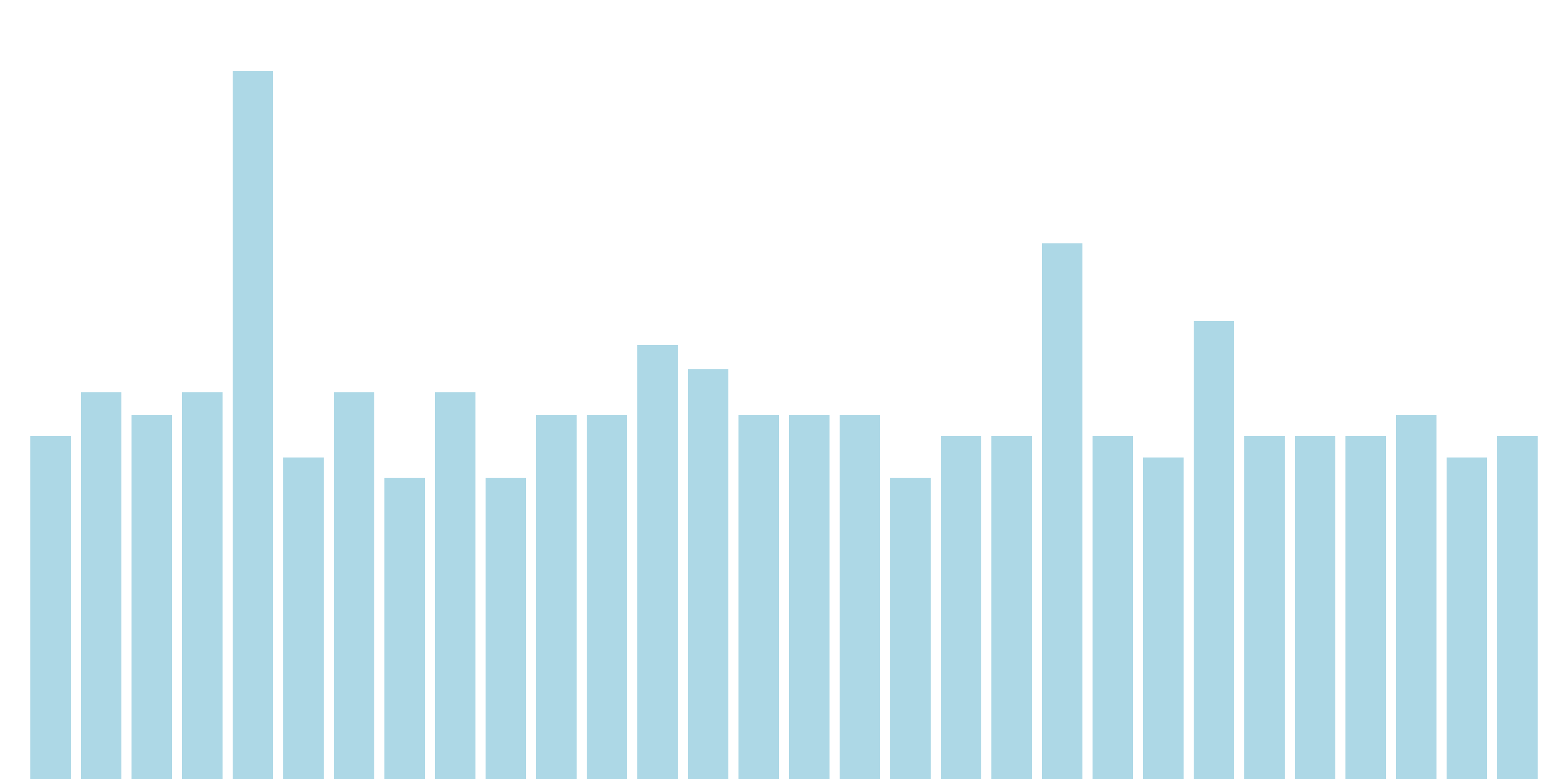}
        \caption{}
    \end{subfigure}
    \hfill
    \begin{subfigure}[b]{0.18\columnwidth}
        \centering
        \includegraphics[width=\textwidth]{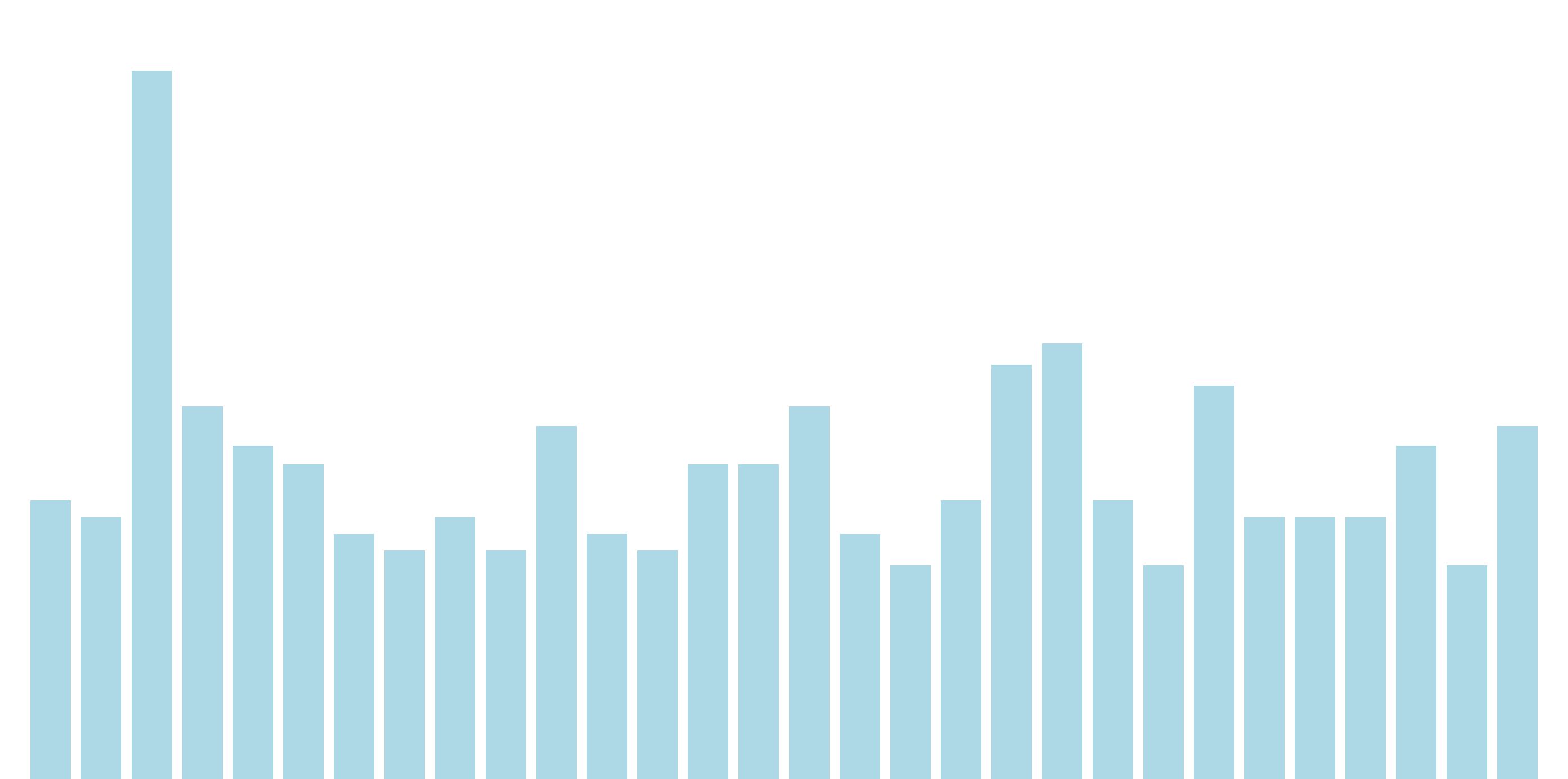}
        \caption{}
    \end{subfigure}
    \hfill
    \begin{subfigure}[b]{0.18\columnwidth}
        \centering
        \includegraphics[width=\textwidth]{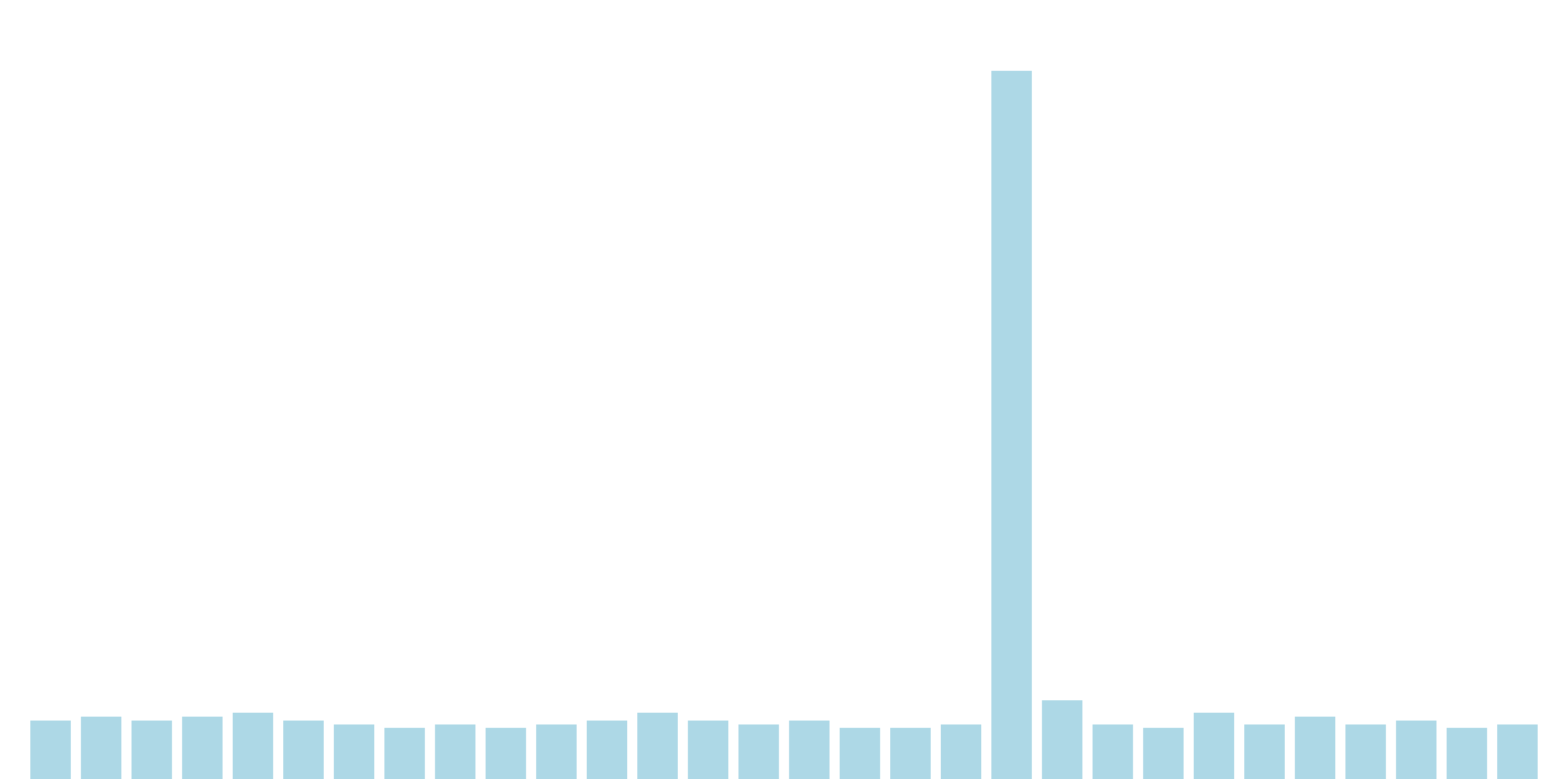}
        \caption{}
    \end{subfigure}
    \\[0.2em]
    \begin{subfigure}[b]{0.18\columnwidth}
        \centering
        \includegraphics[width=\textwidth]{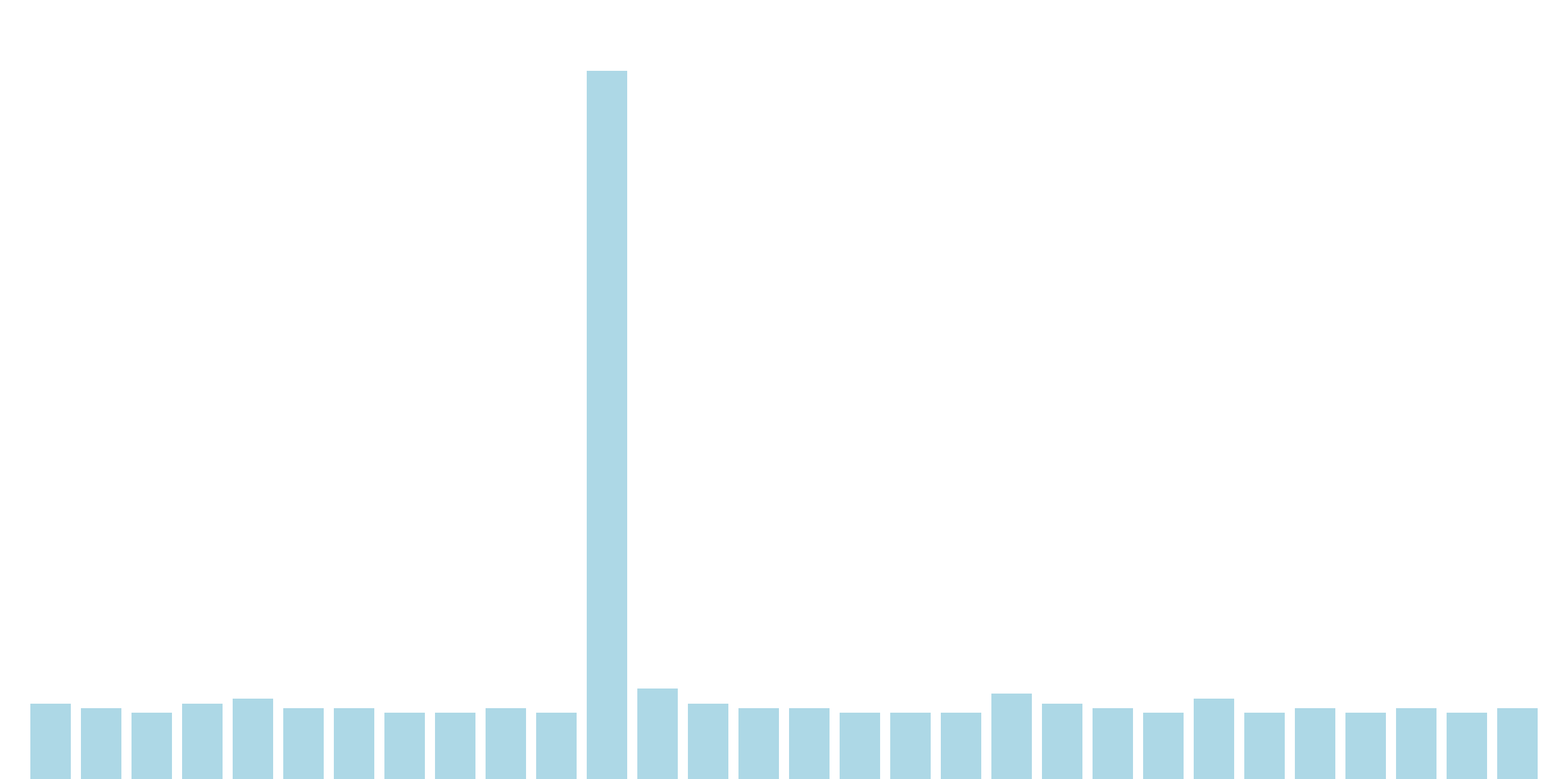}
        \caption{}
    \end{subfigure}
    \hfill
    \begin{subfigure}[b]{0.18\columnwidth}
        \centering
        \includegraphics[width=\textwidth]{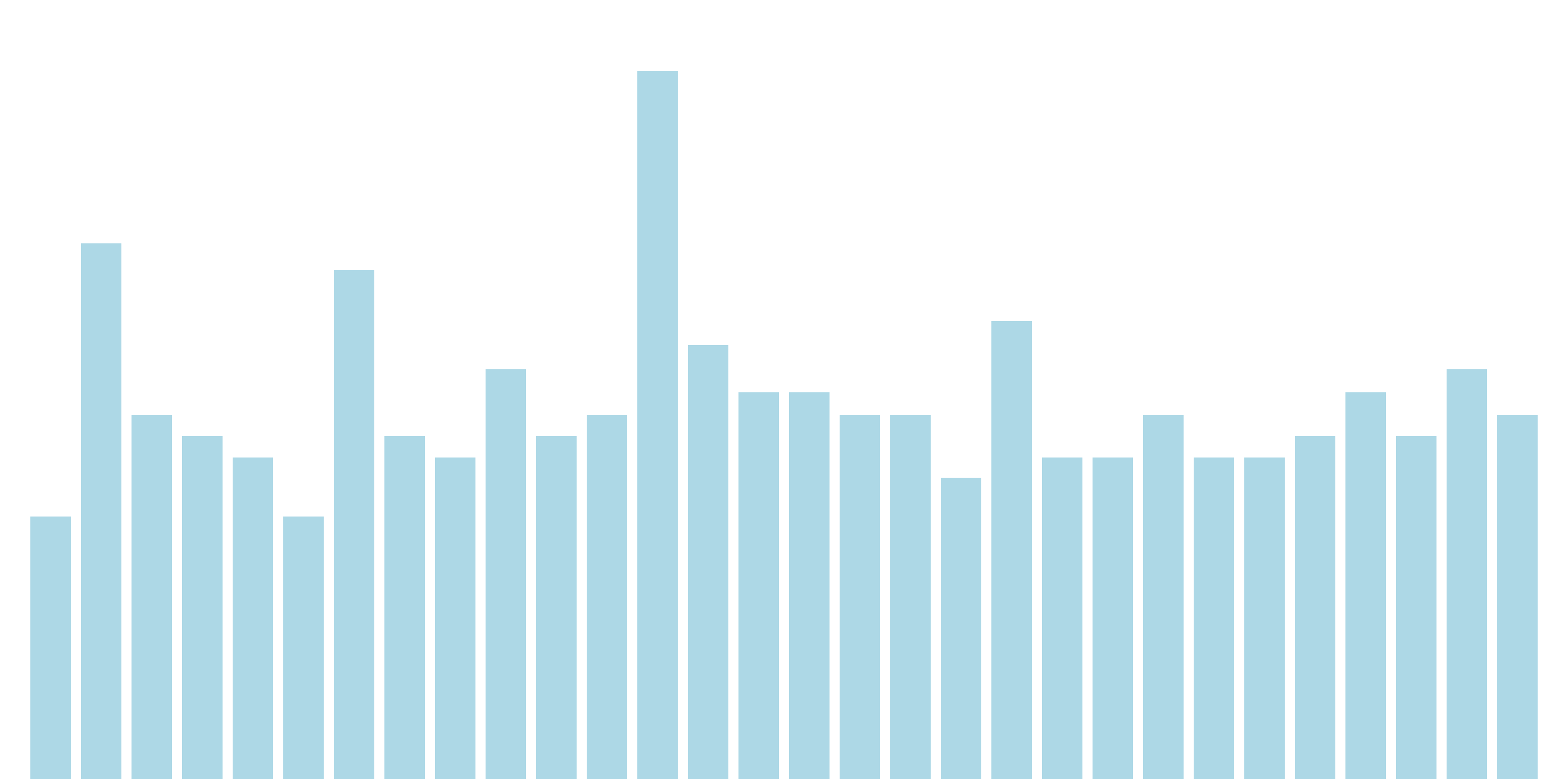}
        \caption{}
    \end{subfigure}
    \hfill
    \begin{subfigure}[b]{0.18\columnwidth}
        \centering
        \includegraphics[width=\textwidth]{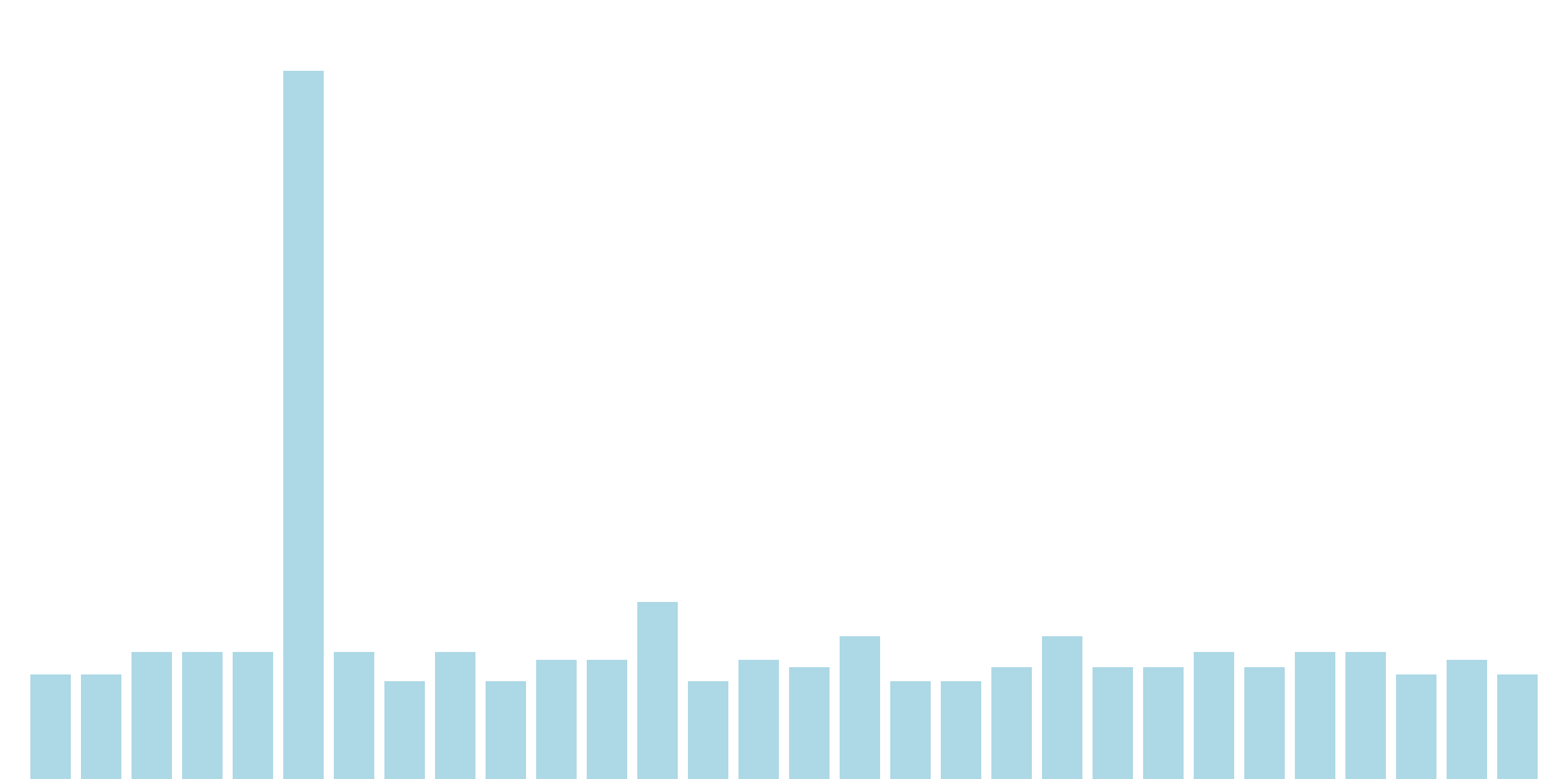}
        \caption{}
    \end{subfigure}
    \hfill
    \begin{subfigure}[b]{0.18\columnwidth}
        \centering
        \includegraphics[width=\textwidth]{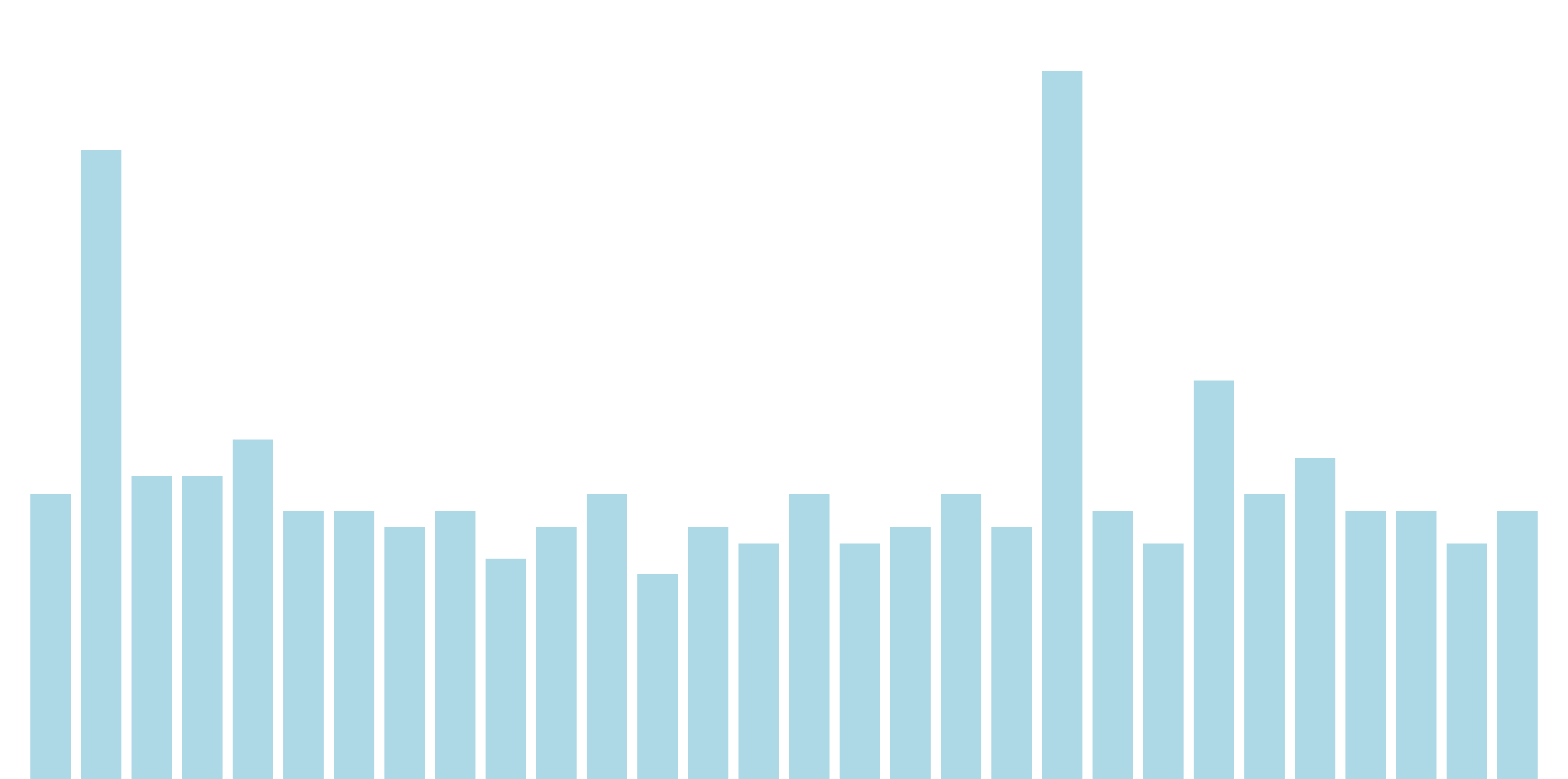}
        \caption{}
    \end{subfigure}
    \\[0.2em]
    \begin{subfigure}[b]{0.18\columnwidth}
        \centering
        \includegraphics[width=\textwidth]{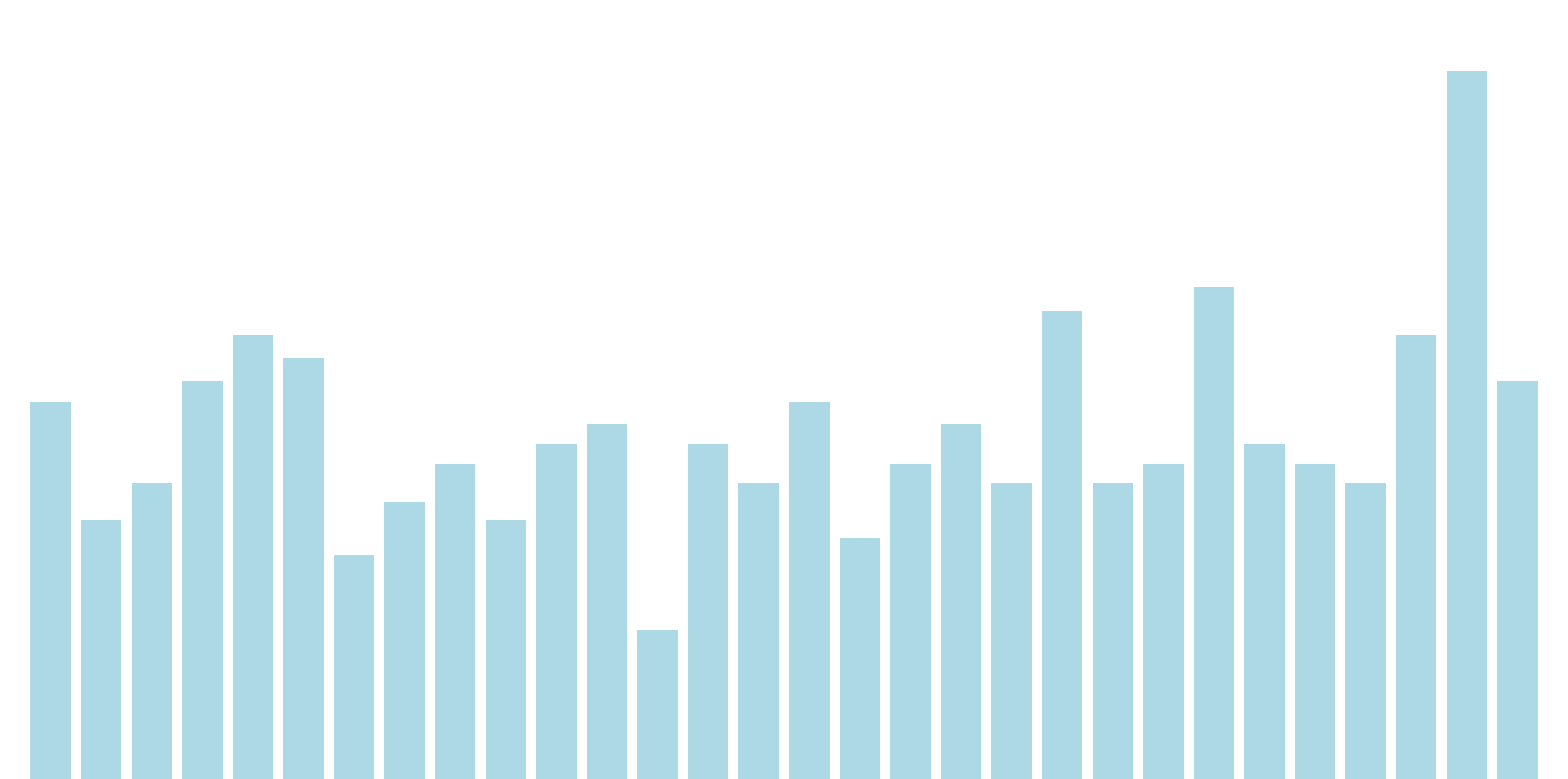}
        \caption{}
    \end{subfigure}
    \hfill
    \begin{subfigure}[b]{0.18\columnwidth}
        \centering
        \includegraphics[width=\textwidth]{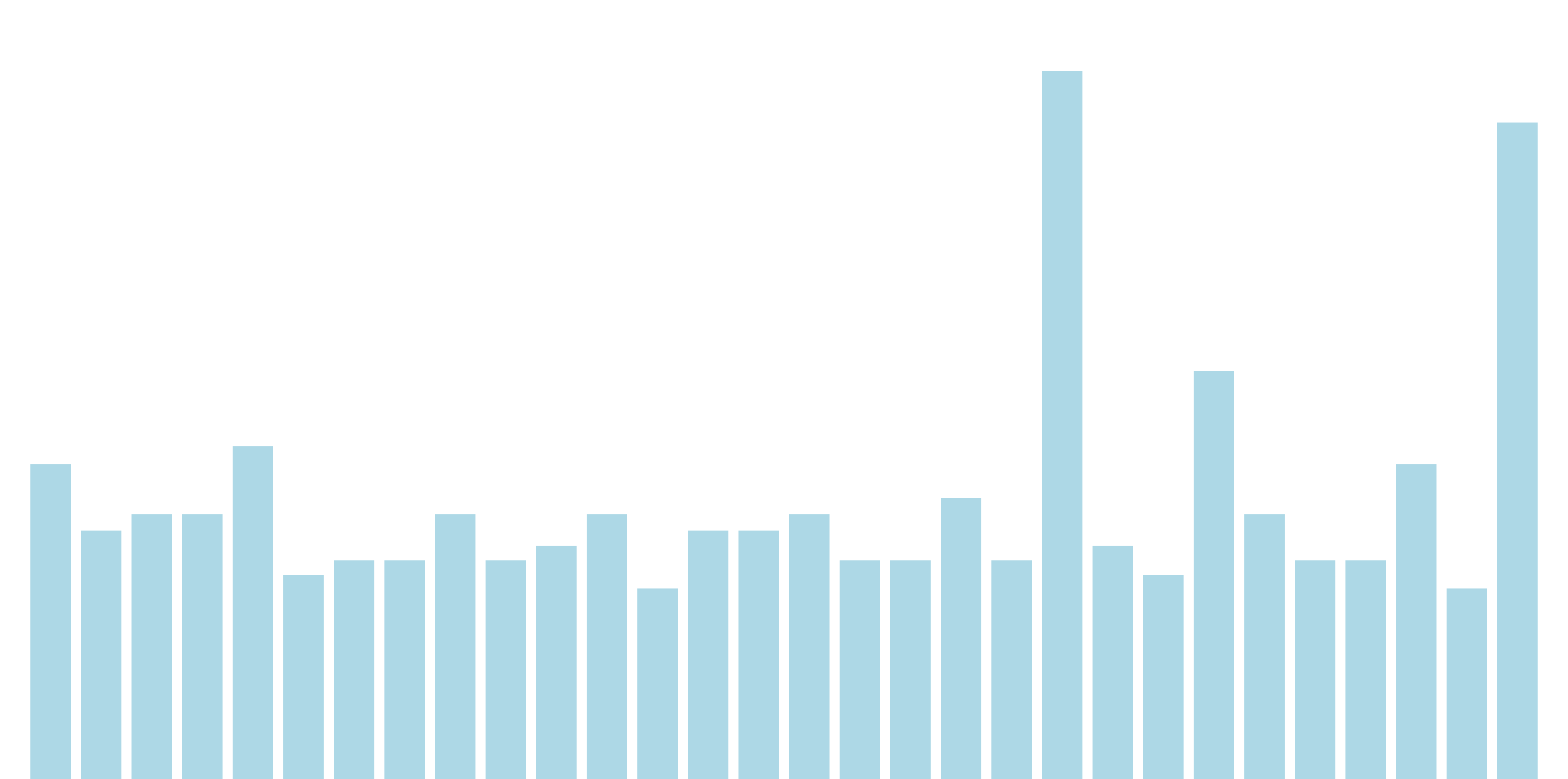}
        \caption{}
    \end{subfigure}
    \hfill
    \begin{subfigure}[b]{0.18\columnwidth}
        \centering
        \includegraphics[width=\textwidth]{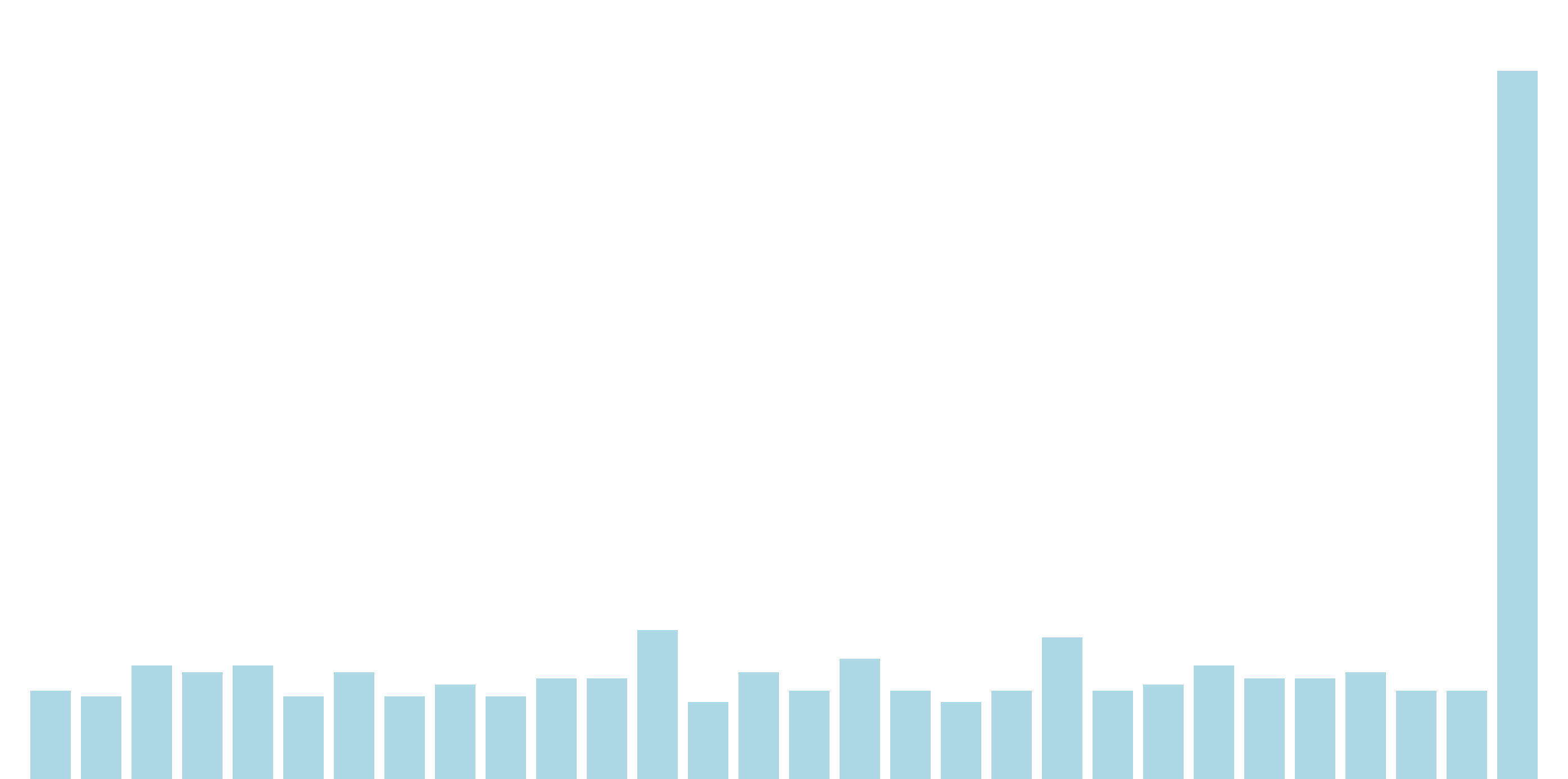}
        \caption{}
    \end{subfigure}
    \hfill
    \begin{subfigure}[b]{0.18\columnwidth}
        \centering
        \includegraphics[width=\textwidth]{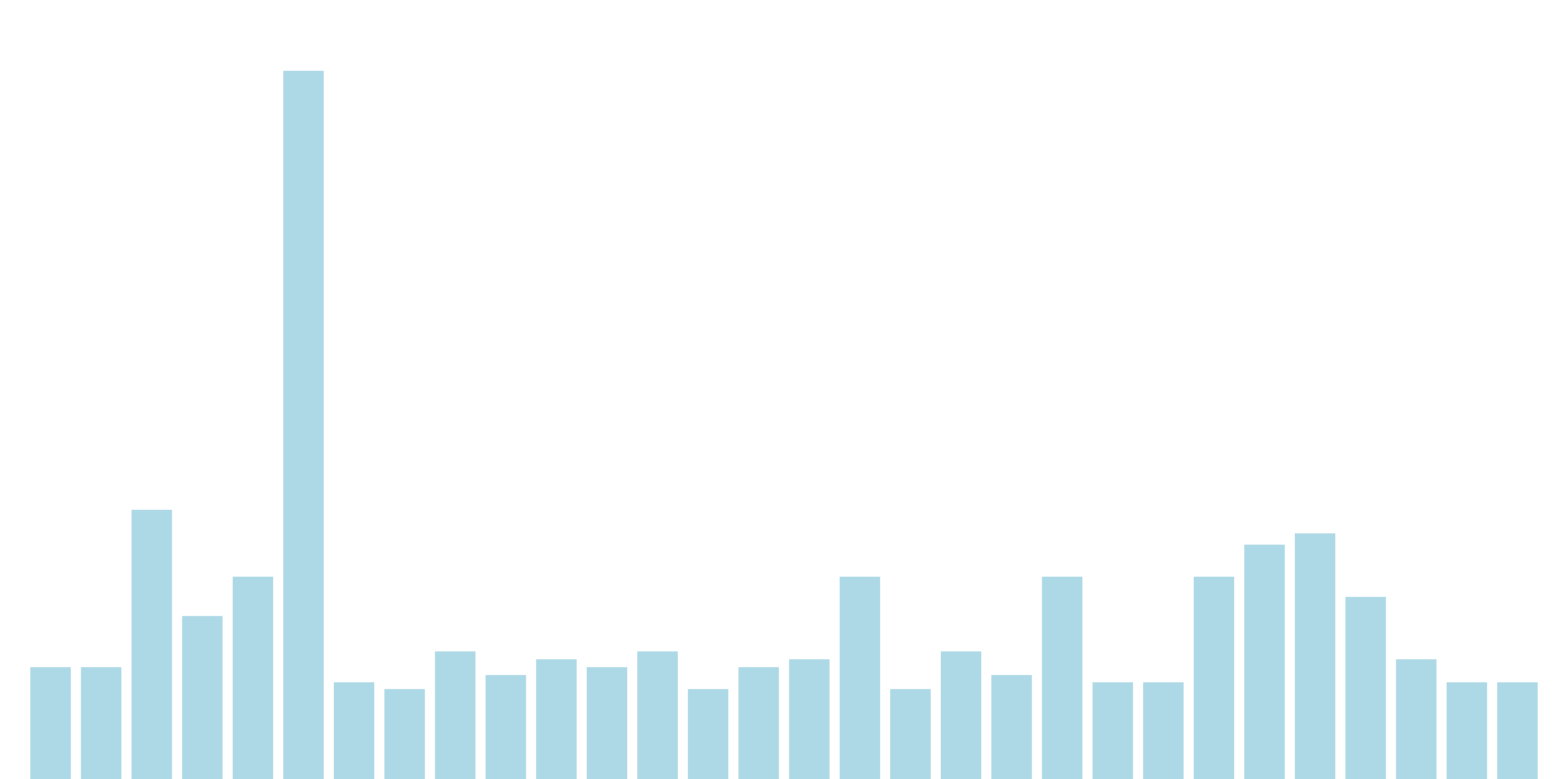}
        \caption{}
    \end{subfigure}
    \\[0.2em]
    \begin{subfigure}[b]{0.18\columnwidth}
        \centering
        \includegraphics[width=\textwidth]{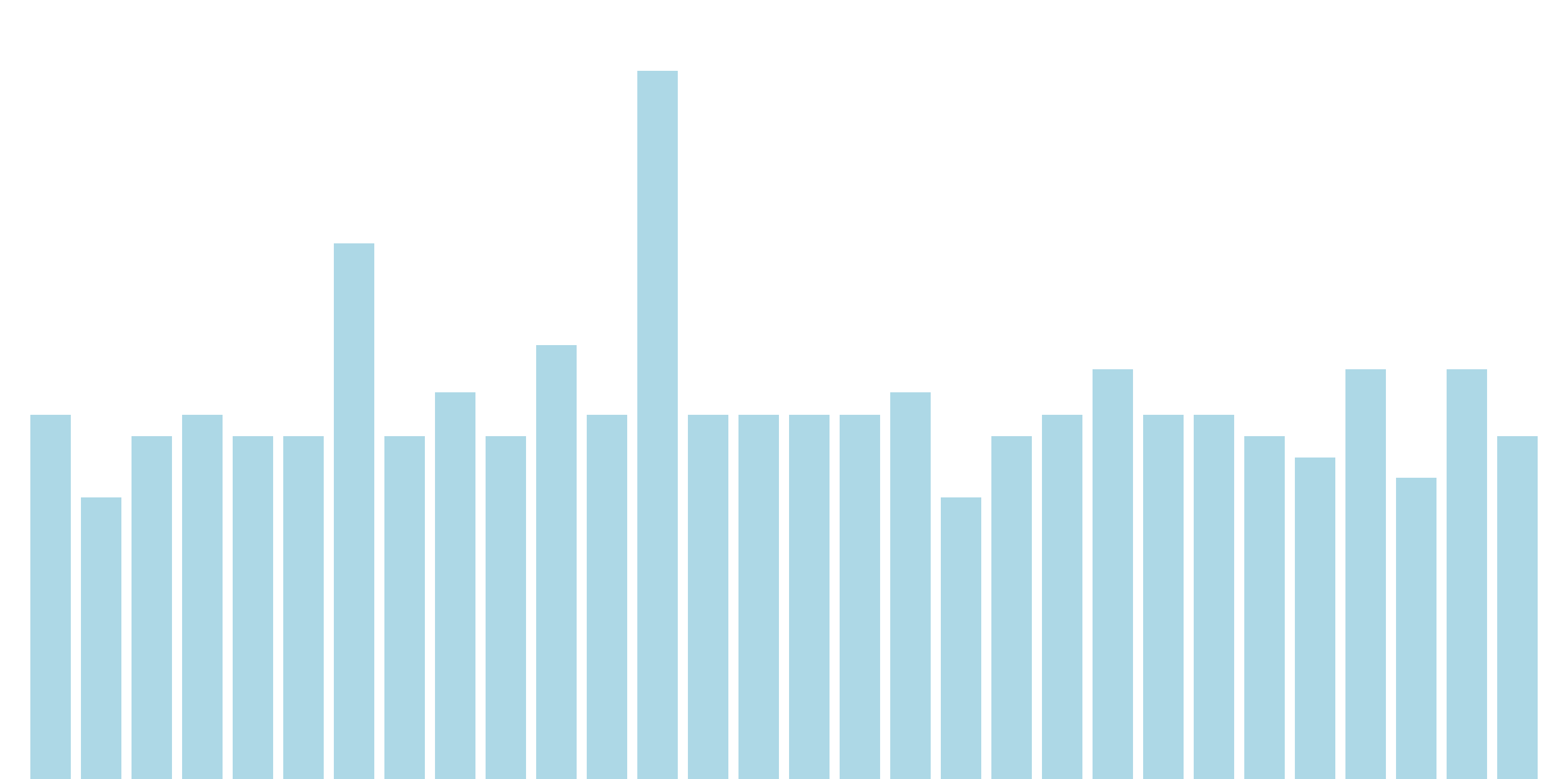}
        \caption{}
    \end{subfigure}
    \hfill
    \begin{subfigure}[b]{0.18\columnwidth}
        \centering
        \includegraphics[width=\textwidth]{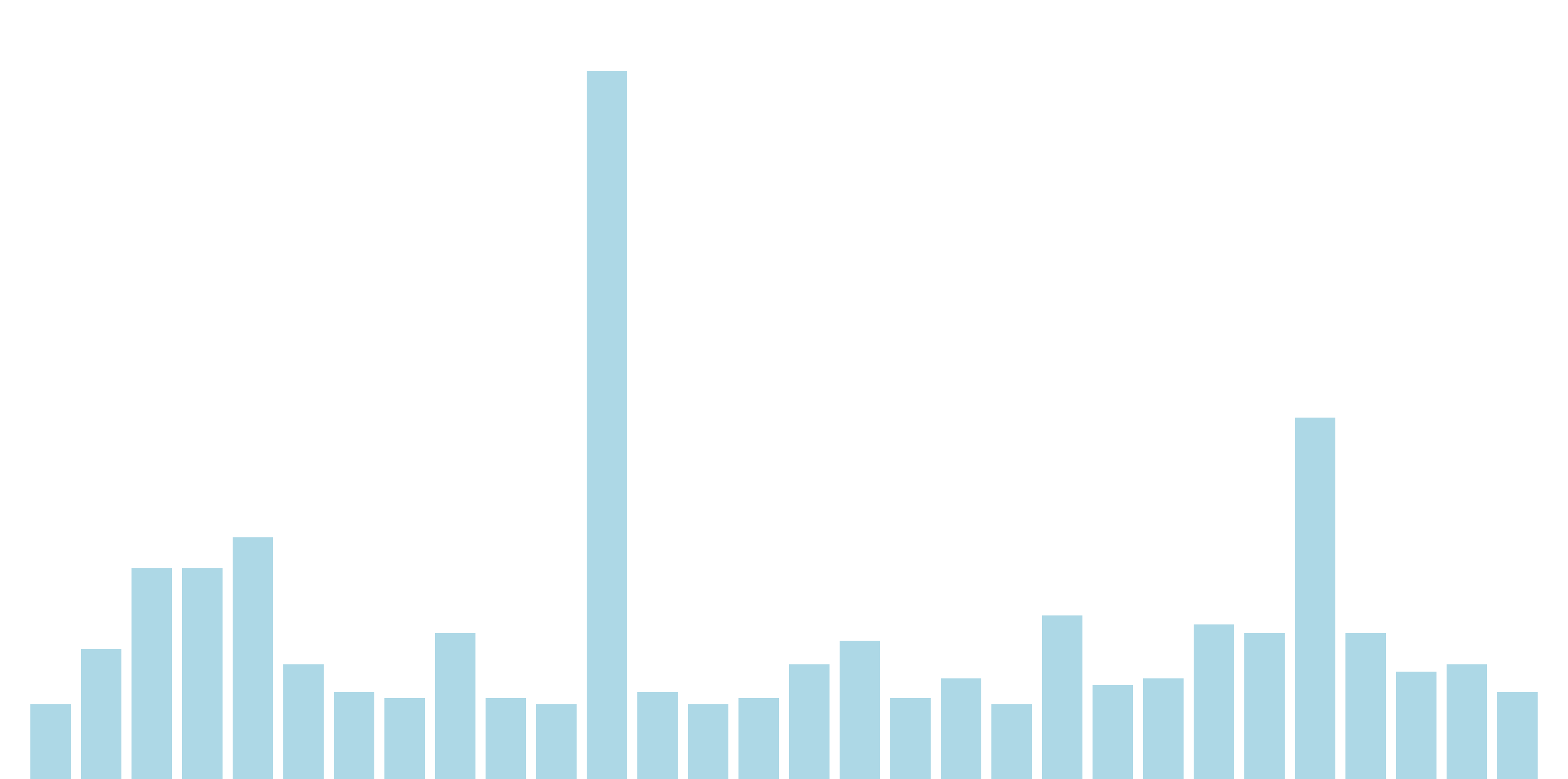}
        \caption{}
    \end{subfigure}
    \hfill
    \begin{subfigure}[b]{0.18\columnwidth}
        \centering
        \includegraphics[width=\textwidth]{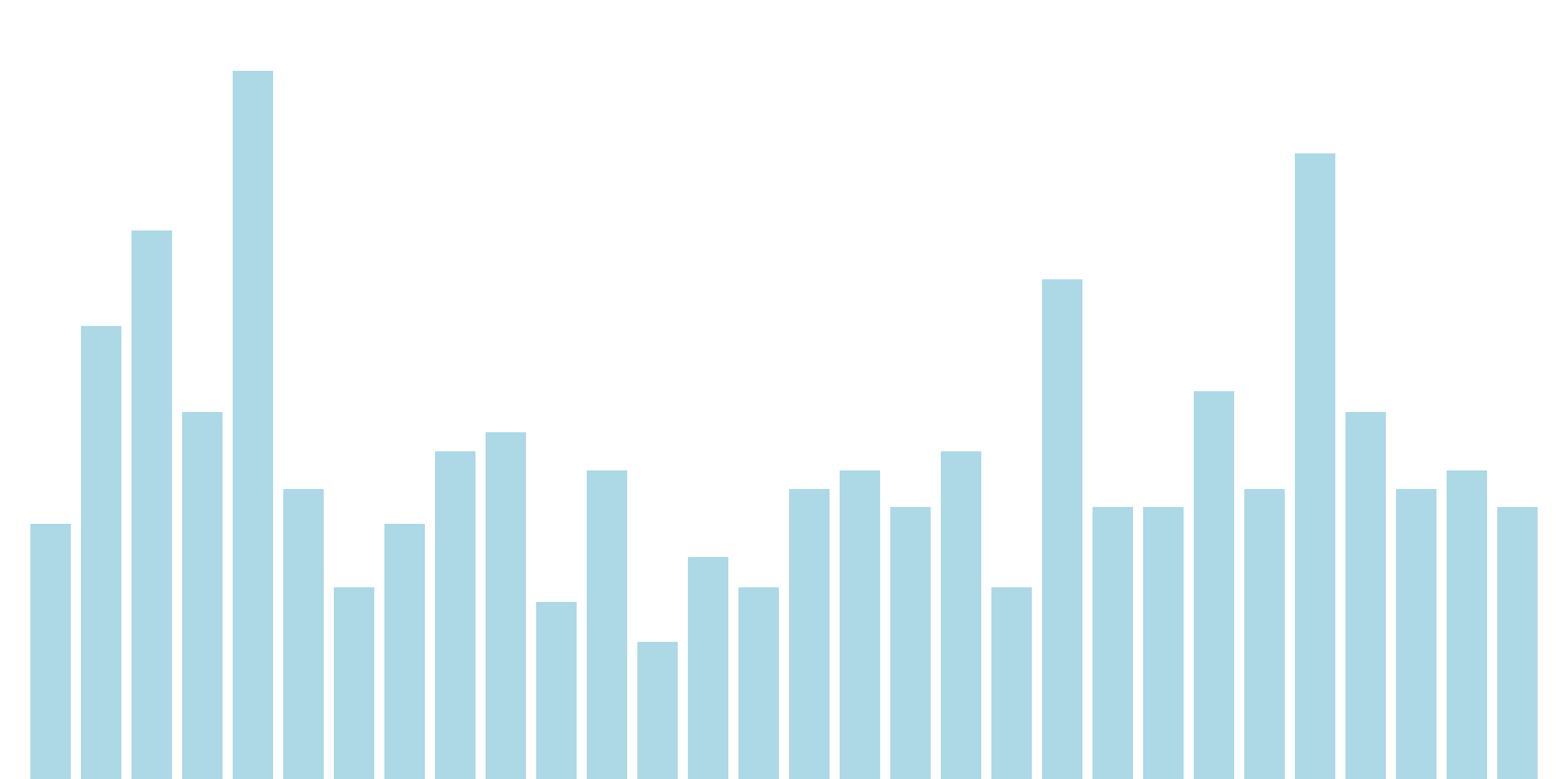}
        \caption{}
    \end{subfigure}
    \hfill
    \begin{subfigure}[b]{0.18\columnwidth}
        \centering
        \includegraphics[width=\textwidth]{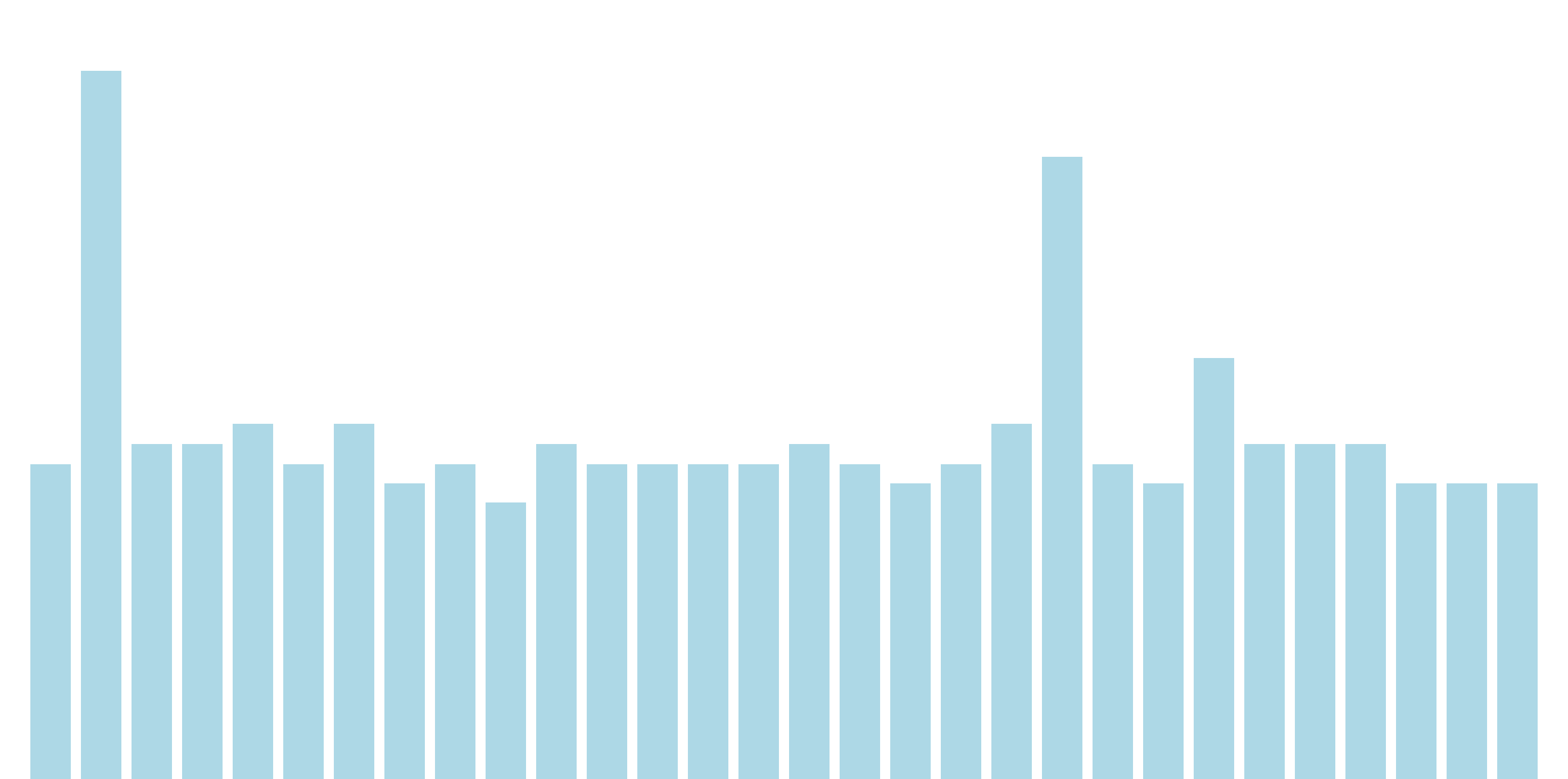}
        \caption{}
    \end{subfigure}
    \caption{Visualization of relative attention weight distributions. Each plot shows how model assigns different importance to local spatial prompts during feature extraction, revealing diverse but effectively complementary spatial-aware learning patterns.}
    \label{fig:weight_dist}
    \vspace{-2pt}
\end{figure}
%-------------------------------------------------------------------------
%-------------------------------------------------------------------------
\subsection{Ablation Study}
We conduct comprehensive ablation studies to analyze each component's contribution, as shown in Table~\ref{tab:ablation}. Starting from vanilla ViT, we observe that basic VPT significantly improves performance (e.g., +17.60\% on CUB200). Our Local Spatial Prompting module shows consistent gains even with a single prompt (LSP*), and the pool variant (LSP) further enhances fine-grained feature capture. When combined with Global Spatial Prompting (GSP), which processes structural information in the frequency domain, our full model achieves optimal performance across all metrics, validating our spatial-domain prompt learning strategy.

%-------------------------------------------------------------------------
\subsection{Visualization Results}
Our analysis revealed that token-dimension saturation leads to local interference between prompts and inefficient feature extraction. We address these through a spatial-aware dual-stream design: local prompts for fine-grained features and global prompts for holistic patterns. The following visualizations demonstrate the effectiveness of our approach.

%-------------------------------------------------------------------------
\textbf{Visualization of Attention Weights:} Figure~\ref{fig:weight_dist} shows the relative attention weight distributions of local spatial prompts across different samples. Notably, these spatial prompts exhibit balanced but distinct attention weights, indicating their complementary roles in capturing different local patterns. The varying distributions further demonstrate our model's ability to adaptively emphasize different spatial regions for effective feature representation.

%-------------------------------------------------------------------------
% 30分支可视化
\begin{figure}[t]
    \centering
    \includegraphics[width=0.9\linewidth]{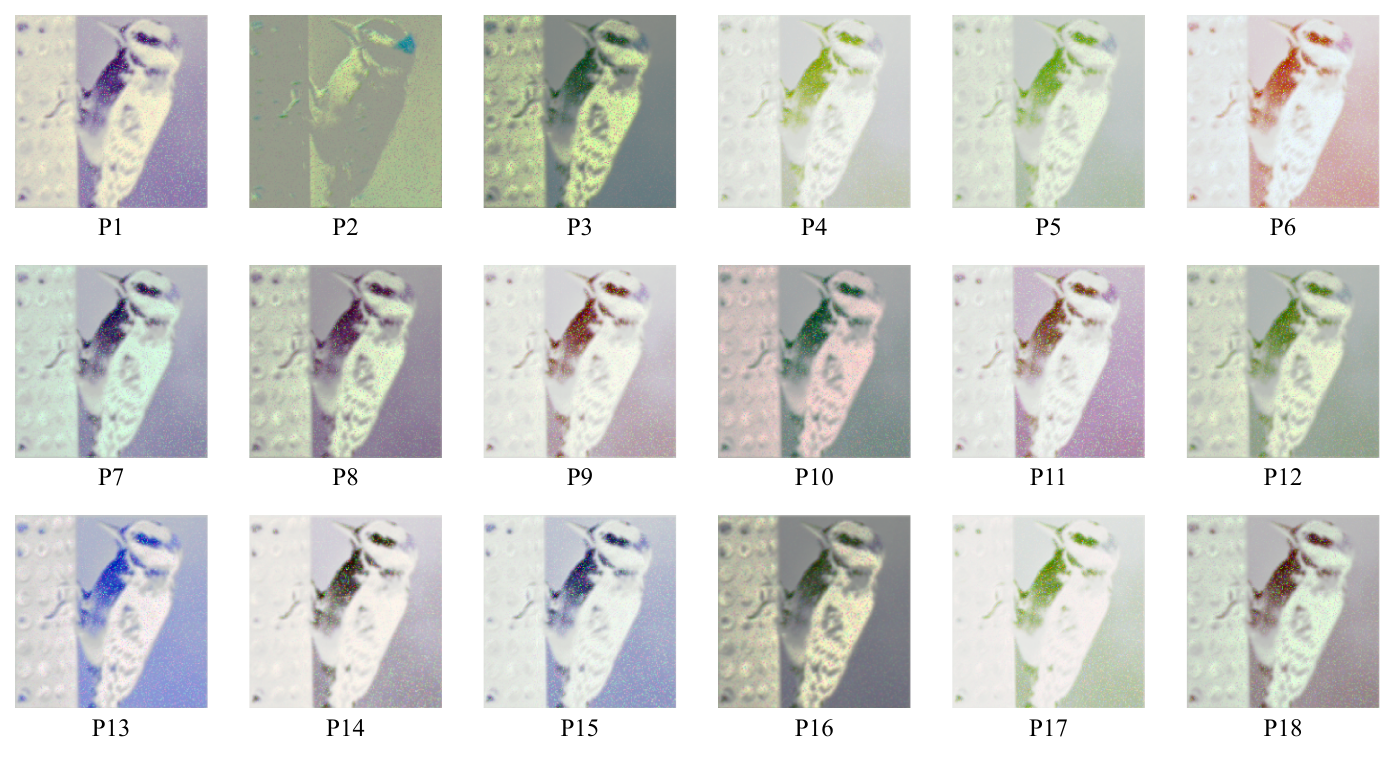}
    \caption{\small Visualization of local spatial prompting (P1-P18): Diverse and complementary patterns across the prompt pool, showing color-independent representation, region-specific focus (e.g., head region in P2-P3, body texture in P7-P8), and multi-scale feature processing for comprehensive fine-grained learning.}
    \label{fig:local_vis}
    \vspace{-1pt}
\end{figure}
%-------------------------------------------------------------------------

\vspace{-3pt}
\textbf{Analysis of Local Spatial Prompts:} Figure~\ref{fig:local_vis} demonstrates the effectiveness of our local spatial prompting mechanism through diverse attention patterns. The prompts exhibit complementary characteristics: (1) \textit{Color-Independent Representation}: encode similar local regions in different color spaces; (2) \textit{Region-Specific Focus}: concentrate on distinct discriminative features (e.g., Prompts 7, 8 on bird's beak); (3) \textit{Multi-Scale Processing}: capture structural and texture details at varying scales. These diverse patterns enable comprehensive fine-grained feature representation without competing for the same attention regions.\vspace{-1pt}

%-------------------------------------------------------------------------
% 频率域可视化
\begin{figure}[t]
    \centering
    % 第一行图片
    \begin{minipage}{\columnwidth}
        \centering
        \includegraphics[width=0.11\columnwidth]{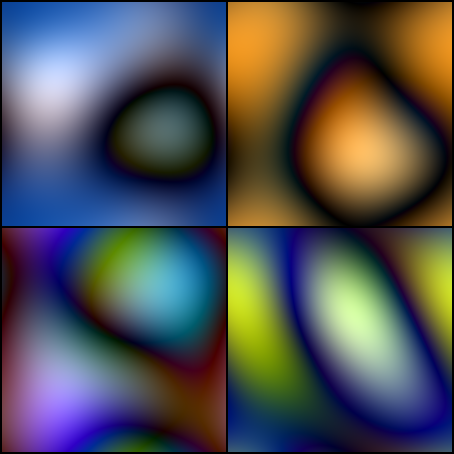}
        \includegraphics[width=0.11\columnwidth]{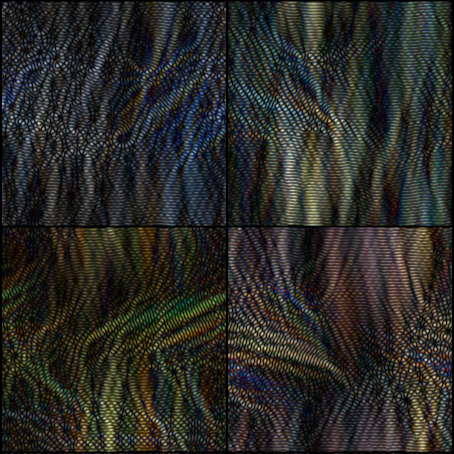}
        \includegraphics[width=0.11\columnwidth]{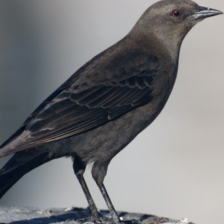}
        \includegraphics[width=0.11\columnwidth]{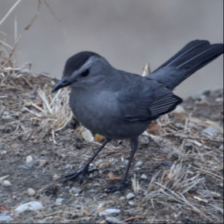}
        \includegraphics[width=0.11\columnwidth]{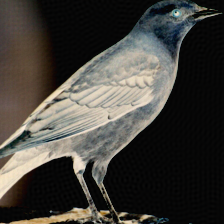}
        \includegraphics[width=0.11\columnwidth]{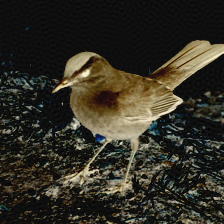}
    \end{minipage}
    
    \vspace{0.3mm}
    % 第二行图片
    \begin{minipage}{\columnwidth}
        \centering
        \includegraphics[width=0.11\columnwidth]{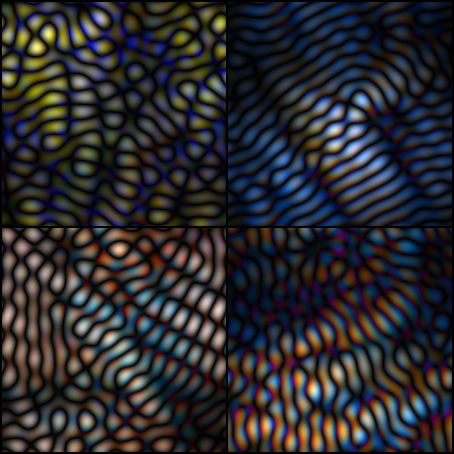}
        \includegraphics[width=0.11\columnwidth]{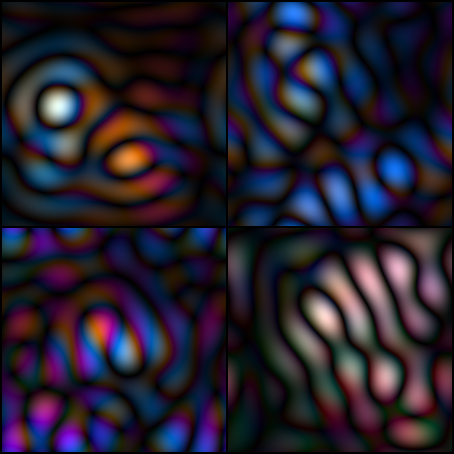}
        \includegraphics[width=0.11\columnwidth]{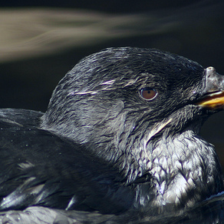}
        \includegraphics[width=0.11\columnwidth]{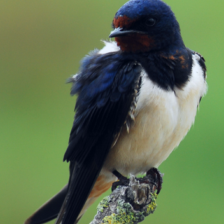}
        \includegraphics[width=0.11\columnwidth]{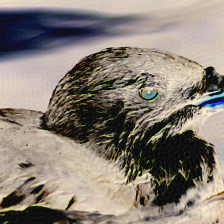}
        \includegraphics[width=0.11\columnwidth]{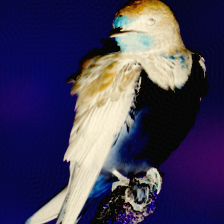}
    \end{minipage}
    
    \vspace{0.3mm}
    % 第三行图片
    \begin{minipage}{\columnwidth}
        \centering
        \includegraphics[width=0.11\columnwidth]{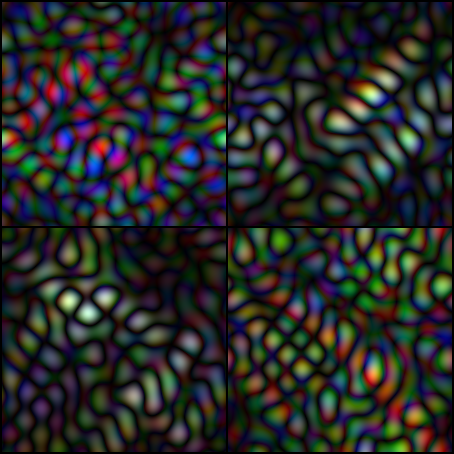}
        \includegraphics[width=0.11\columnwidth]{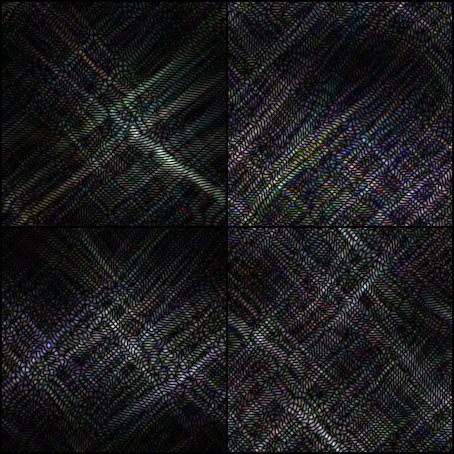}
        \includegraphics[width=0.11\columnwidth]{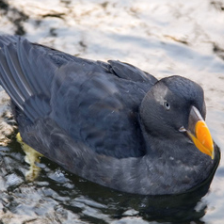}
        \includegraphics[width=0.11\columnwidth]{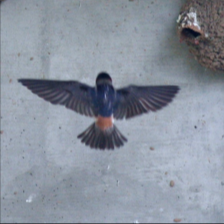}
        \includegraphics[width=0.11\columnwidth]{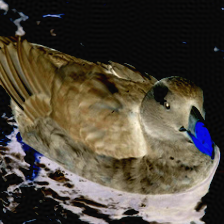}
        \includegraphics[width=0.11\columnwidth]{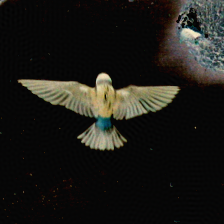}
    \end{minipage}
    
    \vspace{0.3mm}
    % 第四行图片
    \begin{minipage}{\columnwidth}
        \centering
        \includegraphics[width=0.11\columnwidth]{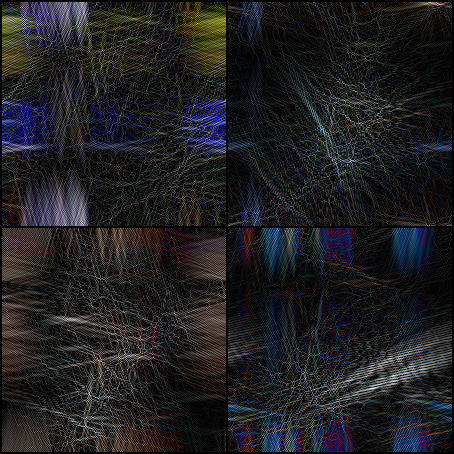}
        \includegraphics[width=0.11\columnwidth]{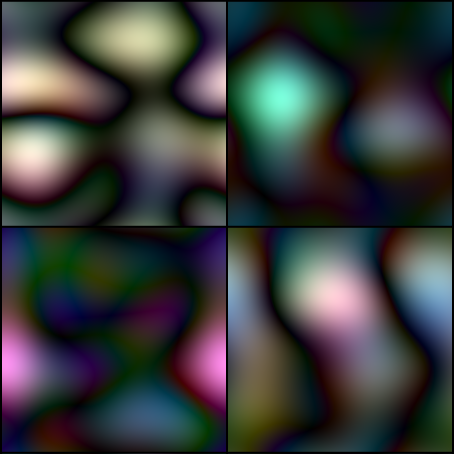}
        \includegraphics[width=0.11\columnwidth]{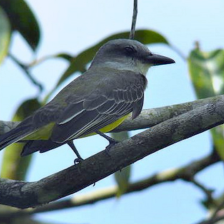}
        \includegraphics[width=0.11\columnwidth]{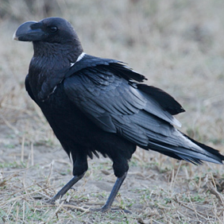}
        \includegraphics[width=0.11\columnwidth]{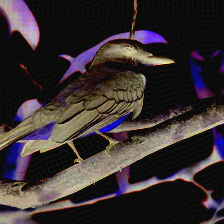}
        \includegraphics[width=0.11\columnwidth]{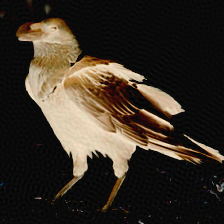}
    \end{minipage}
    
    \vspace{1.2mm}
    {\footnotesize
      \hspace{0.13\columnwidth}(a)\hspace{0.20\columnwidth}(b)\hspace{0.20\columnwidth}(c)\hspace{0.13\columnwidth}%
    }
    \caption{\small Visualization of global spatial prompting: (a) Learned prompts showing hierarchical patterns from holistic structure (top) to fine details (bottom), (b) original images and (c) their spatially-enhanced results, demonstrating emphasis on discriminative features (e.g., wing textures) while preserving global structure.}
    \label{fig:global_vis}
    \vspace{-1pt}
\end{figure}

% before-after热力图
\begin{figure}[t]
    \centering
    % 第一行图片
    \begin{minipage}{\columnwidth}
        \centering
        \includegraphics[width=0.145\columnwidth]{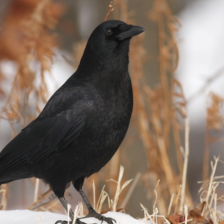}
        \includegraphics[width=0.145\columnwidth]{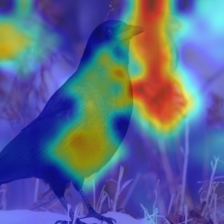}
        \includegraphics[width=0.145\columnwidth]{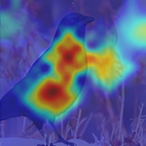}
        \includegraphics[width=0.145\columnwidth]{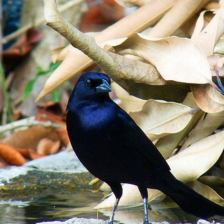}
        \includegraphics[width=0.145\columnwidth]{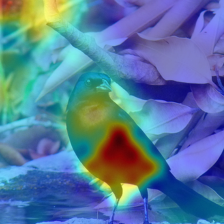}
        \includegraphics[width=0.145\columnwidth]{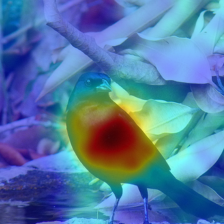}
    \end{minipage}
    
    \vspace{0.4mm}
    % 第二行图片
    \begin{minipage}{\columnwidth}
        \centering
        \includegraphics[width=0.145\columnwidth]{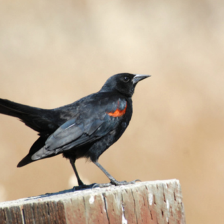}
        \includegraphics[width=0.145\columnwidth]{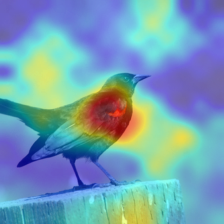}
        \includegraphics[width=0.145\columnwidth]{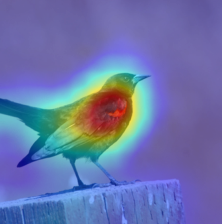}
        \includegraphics[width=0.145\columnwidth]{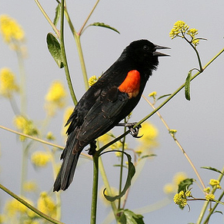}
        \includegraphics[width=0.145\columnwidth]{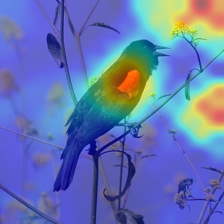}
        \includegraphics[width=0.145\columnwidth]{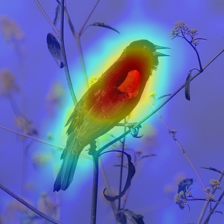}
    \end{minipage}
    
    \vspace{0.4mm}
    % 第三行图片
    \begin{minipage}{\columnwidth}
        \centering
        \includegraphics[width=0.145\columnwidth]{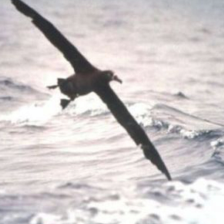}
        \includegraphics[width=0.145\columnwidth]{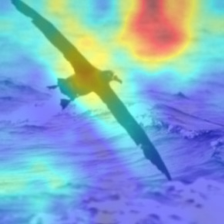}
        \includegraphics[width=0.145\columnwidth]{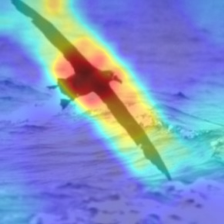}
        \includegraphics[width=0.145\columnwidth]{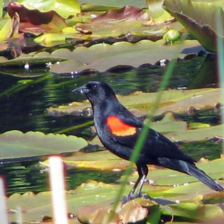}
        \includegraphics[width=0.145\columnwidth]{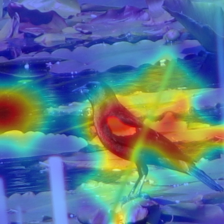}
        \includegraphics[width=0.145\columnwidth]{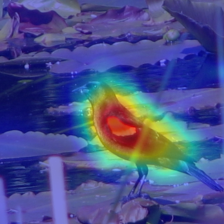}
    \end{minipage}
    
    \vspace{1.5mm}
    {\footnotesize
    \makebox[0.145\columnwidth]{orig}
    \makebox[0.145\columnwidth]{cls}
    \makebox[0.145\columnwidth]{pmt}
    \makebox[0.145\columnwidth]{orig}
    \makebox[0.145\columnwidth]{cls}
    \makebox[0.145\columnwidth]{pmt}
    }
    \caption{\small Comparison of attention patterns: orig shows original images, cls reveals attention scattered on background regions, while pmt demonstrates our framework's ability to focus on discriminative features without token-dimension saturation.}
    \label{fig:attention_maps}
    \vspace{-2pt}
\end{figure}
%-------------------------------------------------------------------------
\textbf{Analysis of Global Spatial Prompts:} Figure~\ref{fig:global_vis}(a) illustrates the learned frequency domain patterns. These prompts demonstrate a hierarchical organization: the top rows show smooth, circular patterns for capturing low-frequency components (holistic structure), while the bottom rows display more intricate patterns for high-frequency details (discriminative textures and edges). This complementary frequency-based representation works alongside local spatial prompts to provide comprehensive structural information, enhancing the model's ability to maintain both global context and fine-grained features.

\textbf{Analysis of Feature Processing:} The comparison between original images (Figure~\ref{fig:global_vis}(b)) and their frequency-enhanced results (Figure~\ref{fig:global_vis}(c)) demonstrates how our global spatial prompting effectively processes structural information. Rather than competing in token dimension, our frequency-based approach adaptively emphasizes discriminative patterns (e.g., wing textures and body shapes) while preserving holistic structural information. This complementary feature processing in frequency domain provides an effective alternative for comprehensive representation.\vspace{-1pt}

\textbf{Analysis of Attention Shift:} Figure~\ref{fig:attention_maps} demonstrates the impact of our LGSP-Prompt framework on attention distribution. Without our spatial prompting mechanisms, the model's attention (cls) primarily focuses on background elements like reeds rather than the foreground bird. After incorporating our dual-perspective framework, the attention map (pmt) shows a clear shift towards discriminative features of the bird itself, demonstrating how our approach effectively guides the model to focus on discriminative features while mitigating token dimension saturation.\vspace{-3pt}

%-------------------------------------------------------------------------
\begin{figure}[t]
    \centering
    \begin{minipage}[t]{0.32\columnwidth}
        \centering
        \includegraphics[height=2cm, width=\linewidth]{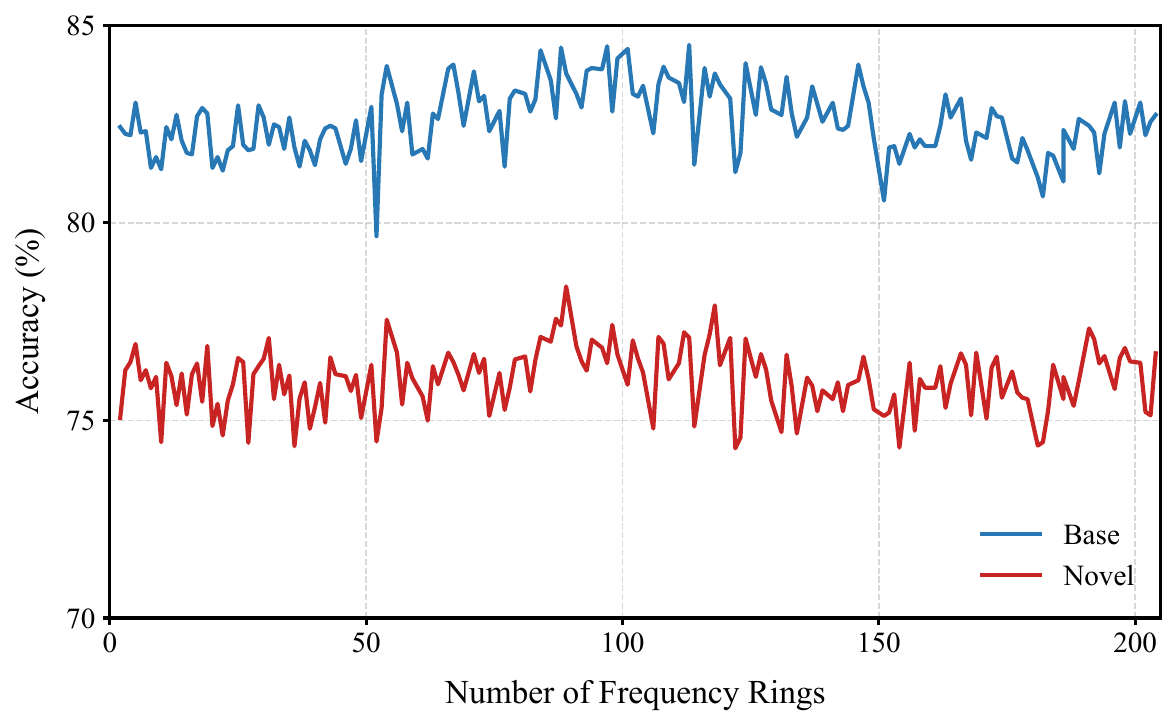}
        \caption*{(a)}
        \label{fig:frequency_rings}
    \end{minipage}
    \hfill
    \begin{minipage}[t]{0.32\columnwidth}
        \centering
        \includegraphics[height=2cm, width=\linewidth]{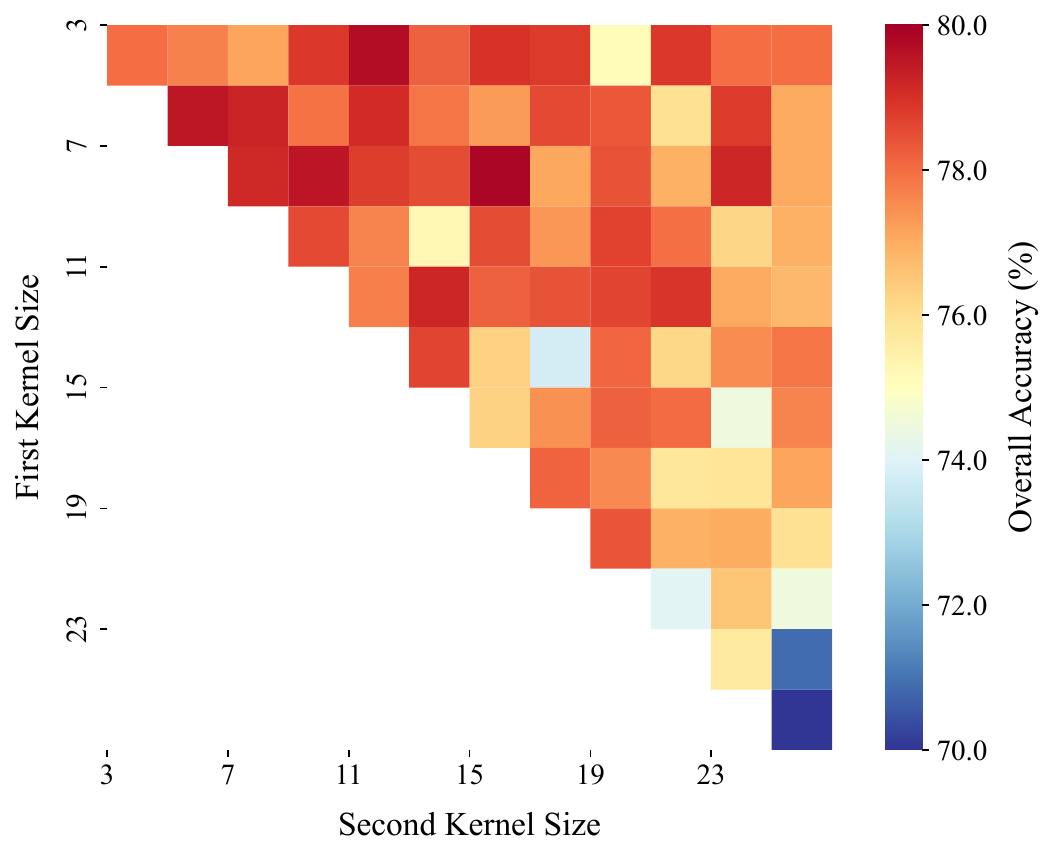}
        \caption*{(b)}
        \label{fig:kernel_performance}
    \end{minipage}
    \hfill
    \begin{minipage}[t]{0.32\columnwidth}
        \centering
        \includegraphics[height=2cm, width=\linewidth]{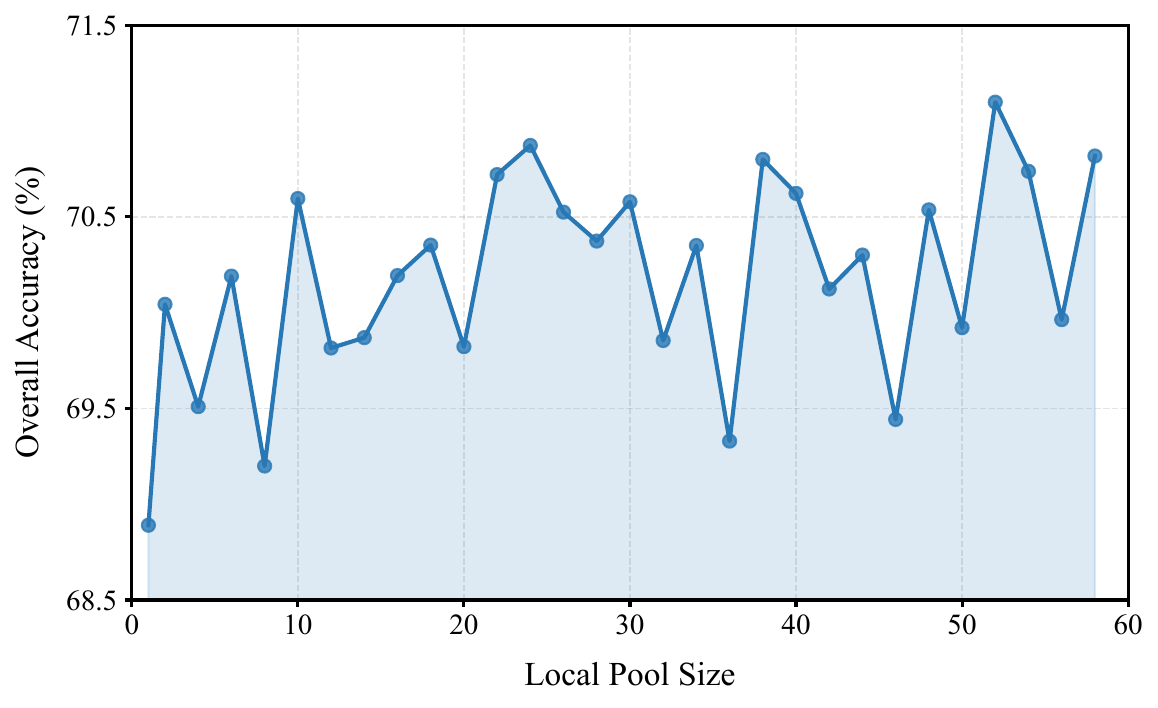}
        \caption*{(c)}
        \label{fig:branch_performance}
    \end{minipage}
    \caption{\small Ablation studies: (a) frequency ring numbers showing robust performance (100 rings optimal), (b) kernel size combinations favoring small-scale processing (1-7), and (c) consistent performance improvement with increased prompt pool capacity.}
    \label{fig:ablation_studies}
    \vspace{-5pt}
\end{figure}
%-------------------------------------------------------------------------
\subsection{Parameter Analysis}
\vspace{-5pt}
\noindent\textbf{Frequency Ring Granularity in GSP:} As shown in Fig.~\ref{fig:ablation_studies}(a), our framework demonstrates remarkable stability across a wide range of frequency ring configurations, with optimal performance around 100 rings (~1.5 pixels/ring). This robustness validates our frequency-domain design principle, where the ring structure effectively segregates and preserves both global structural information in low-frequency bands and discriminative details in high-frequency regions, providing a complementary representation path that avoids token-dimension competition.

\noindent\textbf{Kernel Size Combinations in LSP:} The performance heatmap in Fig.~\ref{fig:ablation_studies}(b) highlights the effectiveness of our multi-scale kernel design. Notably, smaller kernel combinations (1-7) achieve superior performance, aligning with our local-focused design, while larger kernels ($>$11) show degraded results. The complementary interaction between small kernels enables fine-grained feature extraction, effectively capturing discriminative details while mitigating token-dimension saturation in few-shot scenarios.

\noindent\textbf{Local Pool Size:} The analysis of prompt numbers in Local Spatial Pool (Fig.~\ref{fig:ablation_studies}(c)) reveals an overall upward trend despite some fluctuations. The positive trajectory demonstrates our method's ability to effectively leverage increased pool capacity, while maintaining reasonable performance even at smaller pool sizes. This adaptability across different capacities demonstrates that our spatial prompting strategy effectively balances feature diversity and efficiency.\vspace{-3pt}
\vspace*{-3pt}
\section{Conclusion}
\label{sec:con}
\vspace{-3pt}
Our experiments show that Pool and VPT methods face token-dimension conflicts in incremental learning. We propose local-global spatial prompting to address these conflicts while extracting complementary features, achieving superior performance across multiple benchmarks.

\section*{Acknowledgment}
This work is supported by the National Natural Science Foundation of China under grants 62206102; the National Key Research and Development Program of China under grant 2024YFC3307900; the National Natural Science Foundation of China under grants 62436003, 62376103 and 62302184; Major Science and Technology Project of Hubei Province under grant 2024BAA008; Hubei Science and Technology Talent Service Project under grant 2024DJC078; and Ant Group through CCF-Ant Research Fund. The computation is completed in the HPC Platform of Huazhong University of Science and Technology.
{
    \small
    \bibliographystyle{ieeenat_fullname}
    \bibliography{main}
}

% WARNING: do not forget to delete the supplementary pages from your submission 
% \input{sec/X_suppl}

\end{document}